\UseRawInputEncoding

\documentclass[10pt,journal,compsoc]{IEEEtran}

\ifCLASSOPTIONcompsoc
  \usepackage{threeparttable}
  \usepackage[justification=centering]{caption}
  \usepackage[linesnumbered,ruled,vlined]{algorithm2e}
  \usepackage[tbtags]{amsmath}
  \usepackage{epsfig,amssymb,subfigure,bm,dsfont}
  \usepackage{algpseudocode}
  \usepackage{amsfonts}
  \usepackage{array}
  \usepackage{balance}
  \usepackage{bbding}
  \usepackage{booktabs}
  \usepackage{calc}
  \usepackage{cite}
  \usepackage{color}
  \usepackage{curves}
  \usepackage{epstopdf}
  \usepackage{epic}
  \usepackage{float}
  \usepackage{footnote}
  \usepackage{graphicx}
  \usepackage{mdwlist}
  \usepackage{multicol}
  \usepackage{multirow}
  \usepackage{ntheorem}
  \usepackage{pifont}
  \usepackage[english]{babel}
  \usepackage{subfig}
  \usepackage{pifont}

\else
  \usepackage{cite}
\fi

\ifCLASSINFOpdf

\else

\fi

\newcommand{\tabincell}[2]{\begin{tabular}{@{}#1@{}}#2\end{tabular}}
\begin{document}

\newtheorem{definition}{Definition}
\newtheorem{assumption}{Assumption}
\newtheorem{lemma}{Lemma}
\newtheorem{theorem}{Theorem}
\newtheorem{remark}{Remark}
\newtheorem{proposition}{Proposition}
\newtheorem{corollary}{Corollary}

\title{{Blockchain Assisted Decentralized Federated Learning~(BLADE-FL): Performance Analysis and Resource Allocation}}

\author{Jun~Li, Yumeng~Shao, Kang~Wei, Ming~Ding, Chuan~Ma, Long~Shi, Zhu Han, and H.~Vincent~Poor % <-this % stops a space
\IEEEcompsocitemizethanks{\IEEEcompsocthanksitem J. Li, Y. Shao, K. Wei, C. Ma, and L. Shi are with the School of Electronic and Optical Engineering, Nanjing University of Science and Technology, 210094, China. E-mail: \{jun.li, shaoyumeng, kang.wei, chuan.ma\}@njust.edu.cn, slong1007@gmail.com.
\IEEEcompsocthanksitem M. Ding is with Data61, CSIRO, Sydney, Australia. E-mail: ming.ding@data61.csiro.au.
\IEEEcompsocthanksitem Z. Han is with the Department of Electrical and Computer Engineering, University of Houston, Houston, TX, USA. E-mail: hanzhu22@gmail.com.
\IEEEcompsocthanksitem H.~V.~Poor is with Department of Electrical Engineering, Princeton University, NJ, USA. E-mail: poor@princeton.edu.
}
}

\IEEEtitleabstractindextext{%
\begin{abstract}
Federated learning~(FL), as a distributed machine learning paradigm, promotes personal privacy by local data processing at each client. However, relying on a centralized server for model aggregation, standard FL is vulnerable to server malfunctions, untrustworthy server, and external attacks. To address this issue, we propose a decentralized FL framework by integrating blockchain into FL, namely, blockchain assisted decentralized federated learning~(BLADE-FL). In a round of the proposed BLADE-FL, each client broadcasts the trained model to other clients, aggregates its own model with received ones, and then competes to generate a block before its local training of the next round.
%competes to generate a block based on the received models, and then aggregates the models from the generated block before its local training of the next round.
We evaluate the learning performance of BLADE-FL, and develop an upper bound on the global loss function. Then we verify that this bound is convex with respect to the number of overall aggregation rounds $K$, and optimize the computing resource allocation for minimizing the upper bound. We also note that there is a critical problem of training deficiency, caused by lazy clients who plagiarize others' trained models and add artificial noises to disguise their cheating behaviors. Focusing on this problem, we explore the impact of lazy clients on the learning performance of BLADE-FL, and characterize the relationship among the optimal $K$, the learning parameters, and the proportion of lazy clients. Based on MNIST and Fashion-MNIST datasets, we show that the experimental results are consistent with the analytical ones. To be specific, the gap between the developed upper bound and experimental results is lower than $5\%$, and the optimized $K$ based on the upper bound can effectively minimize the loss function.
\end{abstract}

}

\maketitle

\IEEEdisplaynontitleabstractindextext

\IEEEpeerreviewmaketitle

\IEEEraisesectionheading{\section{Introduction}\label{sec:introduction}}

\IEEEPARstart{W}{ith} the development of the Internet of Things~(IoT), the amount of data from end devices is exploding at an unprecedented rate. Conventional machine learning (ML) technologies encounter the problem of how to efficiently collect distributed data from various IoT devices for centralized processing~\cite{TheNextGrandChallenges}. To tackle the issue raised by transmission bottleneck, distributed machine learning~(DML) has emerged to process data at the network edge in a distributive manner~\cite{8805879}. DML can alleviate the burden on the central server by dividing a task into sub-tasks assigned to multiple nodes. However, DML needs to exchange samples when training a task~\cite{DBLP:conf/aistats/McMahanMRHA17}, posing a serious risk of privacy leakage~\cite{9090973}. As such, federated learning (FL)~\cite{DBLP:journals/corr/KonecnyMRR16}, proposed by Google as a novel DML paradigm, shows its potential advantages~\cite{8951246}. In a FL system, a machine learning model is trained across multiple distributed clients with local datasets and then aggregated on a centralized server. FL is able to cooperatively implement machine learning tasks
without raw data transmissions, thereby promoting clients' data privacy~\cite{9048613,DBLP:journals/corr/abs-2007-02056, DBLP:journals/tifs/WeiLDMYFJQP20}. FL has been applied to various data-sensitive scenarios, such as smart health-care, E-commerce~\cite{DBLP:journals/tist/YangLCT19}, and the Google project Gboard~\cite{DBLP:journals/corr/abs-1912-01218}.

However, due to centralized aggregations of models, standard FL is vulnerable to server malfunctions and external attacks, incurring either inaccurate model updates or even training failures. In order to solve this single-point-failure issue, blockchain~\cite{nakamoto2008peer, DBLP:journals/fgcs/ReynaMCSD18,8436042} has been applied to FL systems. Leveraging advantages of blockchain techniques, the work in~\cite{DBLP:journals/corr/abs-1808-03949} developed a blockchain-enabled FL architecture to validate the uploaded parameters and investigated system performance, such as block generation rate and learning latency.
%Later, the work in~\cite{DBLP:conf/cyberc/MartinezFH19} incorporated~\textit{Delegated Proof of Stake}~(DPoS) into blockchain-enabled FL to enhance the delay performance at the expense of robustness.
Most of the works \cite{9347812, 9399813, 9272656}
%, DBLP:journals/access/RahmanHIAM20, 9321132, DBLP:journals/iotj/FanZZC21
have introduced a third-party blockchain network into FL systems, to benefit from a fully decentralized network at the expense of extra mining delay and mining consumption. Since model aggregations are fulfilled by miners in a decentralized manner, the blockchained FL can solve the single-point-failure problem. In addition, owing to a validation process of local training, FL can be extended to untrustworthy devices in a public network~\cite{8470083}.

To protect privacy and security of FL systems, the work in \cite{DBLP:journals/tii/LuHDMZ20a} developed a tamper-proof architecture that utilized blockchain to enhance system security when sharing parameters, and proposed a novel consensus mechanism, i.e., \textit{Proof of Quality}~(PoQ), to optimize reward function. The work in \cite{DBLP:journals/wc/LiuPKINE20} protected privacy of blockchained FL by using local differential privacy, and stand against poisoning attack by executing smart contract. The work in \cite{9347025} applied consortium blockchain and Practical Byzantine Fault Tolerance (PBFT) consensus algorithm into blockchained FL, to ensure the network reliability and safety with a trusted committee. The work in \cite{DBLP:journals/tpds/ShayanFYB21} investigated the robustness of blockchained FL against certain percentage of member attacks, and proposed a corresponding defence mechanism to protect the system. The work in \cite{DBLP:conf/wcnc/Pokhrel020} proposed an autonomous blockchain based FL design for privacy-aware and efficient vehicular communication network, and achieved the system-level performance by adjusting parameters such as block size, block arrival rate, and retransmission limit.

Later, other works have investigated different directions of blockchain assisted FL. For example, the work in \cite{DBLP:journals/access/ToyodaZZM20} proposed an incentive mechanism to encourage clients to follow the protocol of  blockchained FL. The recent work in \cite{DBLP:journals/network/LuHZMZ21} modeled the computation and communication resource in the blockchained FL, and improved the utility between learning performance and resource consumption by controlling the number of local iterations in FL. The work in \cite{9357330} showed the energy consumption model, and optimized the performance by allocating energy resources. The work in~ \cite{DBLP:journals/iotj/ZhangLYLLLCXZ21} applied blockchained FL into industrial IoT for device failure detection, so that the system can be applied in reality.

Although the above mentioned works resorted to blockchain architecture for avoiding single-point-failure, they inevitably introduced a third-party, i.e., miners rooted from blockchain, to store the aggregated models distributively, causing potential information leakage. Also, these works did not analyze the convergence performance of model training, which is important for evaluating FL learning performance. In addition, the consumption of resources, e.g., computing capability, caused by mining in blockchain~\cite{8946151} is generally not taken into account in these works. However, resources consumed by mining are not negligible compared with those consumed by FL model training~\cite{DBLP:journals/fgcs/ReynaMCSD18}, especially when mining with mobile phones has become a reality with the development of user terminals~(e.g., mining for android.). Hence, blockchain-enabled FL needs to balance computational resource allocation between training and mining.

In this work, we propose a novel blockchain assisted decentralized FL~(BLADE-FL) architecture. In our BLADE-FL, training and mining processes are incorporated and implemented at each client, i.e., a client conducts both training and mining tasks with its own computing capability. The conventional framework that FL and blockchain are separated and respectively executed by training clients and miners can be deemed as a special case of resource allocation in our BLADE-FL. That is, in the conventional framework, a part of clients allocate all the computation resources for training, and the other part for mining only.
%In more detail, the participant clients can be grouped into two disjoint sets, i.e., one performing FL only and the other one conducting mining only.

To be specific, we analyze an upper bound on the loss function to evaluate the learning performance of BLADE-FL. Then we optimize the computing resource allocation between local training and mining on a client to approach optimal learning performance. We also pay special attentions to a security issue that inherently exists in BLADE-FL, known as \emph{lazy clients} problem. In this problem, lazy clients try to save their computing resources by directly plagiarizing models from others, leading to training deficiency and performance degradation. In this case, we explore the impact of lazy clients on the learning performance, and optimize $K$ for minimizing the loss function.

The main contributions can be summarized as follows.

\begin{itemize}
        \item {We propose a novel blockchain-assisted FL framework, namely, BLADE-FL, to overcome the issues raised by centralized aggregations in conventional FL systems.
            %In each round of BLADE-FL, clients first train local models and broadcast them to others. Then they play the role as miners to compete for generating a block based on the received models. Afterwards, each client aggregates these models from the verified block to form an initial model utilized for the local training in the next round.
            Compared with conventional blockchain-enabled FL, our BLADE-FL helps promote privacy against model leakage, and guarantees tamper-resistant model updates in a trusted blockchain network.}
        \item {We analyze an upper bound on the loss function to evaluate the learning performance of BLADE-FL. In particular, we minimize the upper bound by optimizing the computing resource allocation between training and mining, and further explore the relationship among the optimal number of integrated rounds, the training time per iteration, the mining time per block, the number of clients, and the learning rate.}
        \item{We focus on a lazy model for BLADE-FL, where the lazy clients plagiarize others' weights and add artificial noises. Moreover, we develop an upper bound on the loss function for this case, and investigate the impact of the number of lazy clients and the power of artificial noises on the learning performance.}
        \item{We provide experimental results, which are consistent with analytical results. In particular, the developed upper bound on the loss function is tight with reference to the experimental ones (e.g., the gap can be lower than $5\%$), and the optimized resource allocation approaches the minimum of the loss function.
            }
    \end{itemize}

\begin{table}[t]
\caption{Summary of main notation}
\centering
\begin{tabular}{l||l}
\hline
Notation & Description\\
\hline\hline
\hline
$D_i$& The set of training samples in the $i$-th client\\
\hline
$C_i$& The $i$-th client\\
\hline
$N$& The total number of clients\\
\hline
$M$& The total number of lazy clients\\
\hline
\multirow{2}*{$\sigma^2$}& The variance of artificial noise added by\\
& lazy clients\\
\hline
$K$& The total number of integrated rounds\\
\hline
$\tau$ & The number of iterations of local training\\
\hline
$F(\bar{\boldsymbol{w}})$& The global loss function\\
\hline
$F_{i}(\boldsymbol{w})$& The local loss function of the $i$-th client\\
\hline
\multirow{2}*{$\boldsymbol{w}_{i}^{k}$} & Local model weights of the $i$-th client\\
& at the $k$-th integrated round\\
\hline
\multirow{2}*{$\bar{\boldsymbol{w}}^{k}$} & Global model weights aggregated from local\\
&models at the $k$-th integrated round\\
\hline
\multirow{2}*{$\tilde{\boldsymbol{w}}_{i'}^{k}$} & Local model weights of the $i'$-th lazy client\\
& at the $k$-th integrated round\\
\hline
$\eta$& Learning rate of gradient descent algorithm \\
\hline
$\alpha$& Training time per iteration\\
\hline
$\beta$& Mining time per block\\
\hline
$t^{\mathrm{sum}}$& Total computing time constraint of a FL task\\
\hline
\end{tabular}
%\vspace{-0.2cm}
\label{tab:summ_nota}
\end{table}

The remainder of this paper is organized as follows.
Section~\ref{sec:Rela_Work} first introduces the background of this paper.
%The background is introduced in Section~\ref{sec:Rela_Work}.
%We summarize related works in Section~\ref{sec:related}.
Then we propose BLADE-FL in Section~\ref{sec:System model}, and optimize the upper bound on the loss function in Section~\ref{sec:Con_FL}.
Section \ref{Sec:Lazy} investigates the issue of lazy clients.
Section \ref{Sec:privacy} discusses the privacy issue.
The experimental results are presented in Section~\ref{sec:Exm_Res}.
Section~\ref{sec:Concl} concludes this paper.
In addition, Table~\ref{tab:summ_nota} lists the main notations used in this paper.

%=======================================================================================================

%\section{Related Work}\label{sec:related}

\section{Preliminaries}\label{sec:Rela_Work}
\subsection{Federated Learning}
In a FL system, there are $N$ clients with the $i$-th client possessing the dataset $D_i$ of size $\vert D_i \vert$, $i=1,2,\dots,N$.  Each client trains its local model, e.g., a deep neural network, based on its local data and transmits the trained model to the server. Upon receiving the weights from all the clients, the server performs a global model aggregation. There are a number of communication rounds for exchanging models between the server and clients. Each round consists of an uploading phase where the clients upload their local models, and a downloading phase where the server aggregates the model and broadcasts it to the clients. Clients then update their local models based on the global one.

In the $k$-th communication round, the server performs a global aggregation according to some combining rule, e.g., $\bar{\boldsymbol{w}}^{k}=\frac{1}{N}\sum_{i=1}^N \boldsymbol{w}_{i}^{k}$, where $\boldsymbol{w}_{i}^{k}$ and $\bar{\boldsymbol{w}}^{k}$ denote the local weights of the $i$-th client and the aggregated weights, respectively. The global loss function is defined as
$F(\bar{\boldsymbol{w}}^k) = \frac{1}{N}\sum_{i=1}^N F_i(\bar{\boldsymbol{w}}^k)$~\cite{DBLP:journals/corr/abs-1907-09693},
where $F_i(\cdot)$ is the local loss function of the $i$-th client.
In FL, each client is trained locally to minimize the local loss function, while the entire system is trained to minimize the global loss function $F(\bar{\boldsymbol{w}}^k)$.
The FL system finally outputs $F(\bar{\boldsymbol{w}}^K)$, where $K$ is the overall communication rounds. Different from the training process in conventional DML systems~\cite{DBLP:conf/aistats/McMahanMRHA17}, each client in FL only shares their local models rather than their personal data, to update the global model, promoting the clients' privacy.
\subsection{Blockchain}
Blockchain is a shared and decentralized ledger.
%In a blockchain system, each block stores a group of transactions. The blocks are linked together to form a chain by referencing the hash value of the previous block. Owing to the cryptographic chain structure, any data modification within any block destroys the preceding chain structure. Therefore, it is impossible to tamper with the data that has been stored in the blockchain.
Thanks to the consensus mechanism, each transaction included in the newly generated block is immutable. The consensus mechanism validates the data within the blocks and ensures that all the nodes participating in the blockchain store the same data. The most prevalent consensus mechanism is \emph{Proof of Work}~(PoW), used in the Bitcoin system~\cite{nakamoto2008peer}.

In Bitcoin, the process of the block generation is as follows. First, a node broadcasts a transaction with its signature to the blockchain network by the gossip protocol~\cite{DBLP:journals/sigops/DemersGHILSSST88}. Then the nodes in blockchain verify the transaction by the signature.
Afterward, each node collects the verified transactions and competes to generate a new block that includes these transactions, by finding a unique, one-time number~(called a \emph{nonce}). This is to make the hash value of the data meet a specific target value. The node that finds the proper nonce is eligible to generate a new block and broadcasts the block to the entire network. Finally, the nodes validate the new block and append the verified block into the existing blockchain~\cite{DBLP:journals/cem/PuthalMMKD18}.

Notably, the work in PoW is a mathematical problem that is easy to verify but extremely hard to solve. The nodes in the blockchain consume massive computing resources to figure out this complex problem. This process is called \emph{mining}, and those who take part in it are known as miners. Because of the mining process, PoW can defense attacks on the condition that the total computing power of malicious devices are less than the sum of honest devices~(i.e., 51\% attack)~\cite{DBLP:journals/cacm/EyalS18}.

%In this context, blockchain is safe and reliable with the aid of the consensus mechanism. Driven by these merits, we deploy the blockchain to replace the central server, and build up a decentralized FL network with privacy protection.

\section{Proposed Framework}\label{sec:System model}
In this section, we detail the proposed BLADE-FL framework in Section \ref{subsec_framework} and develop a computing resource allocation model in Section \ref{subsec_CRA}.
\subsection{BLADE-FL}\label{subsec_framework}
Our BLADE-FL system consists of $N$ clients each with equal computing power~(the computing power is measured by CPU cycles per second). In this distributed system, each client acts as not only a trainer but also a miner, and the role transition is designed as follows. First, each client (as a trainer) trains the local model, and then broadcasts the local model to the entire network as a requested transaction of the blockchain. Second, the client~(as a miner) mines the block that includes all the local models that are ready for aggregation. Once the newly generated block is validated by the majority of clients, the verified models in the block are immutable. Without the intervention of any centralized server, each client performs the global aggregation to update its local model by using all the shared models in the validated block. Suppose that the uploading and downloading phases cannot be tampered with external attackers.

\begin{figure}[t]%%%%%%%%%%%%%%%%%½ûֹͼƬ¸¡¶¯
  \centering
  % Requires \usepackage{float}À´½ûÖ¹¸¡¶¯
  \includegraphics[width=0.5\textwidth,height=0.6\textwidth]{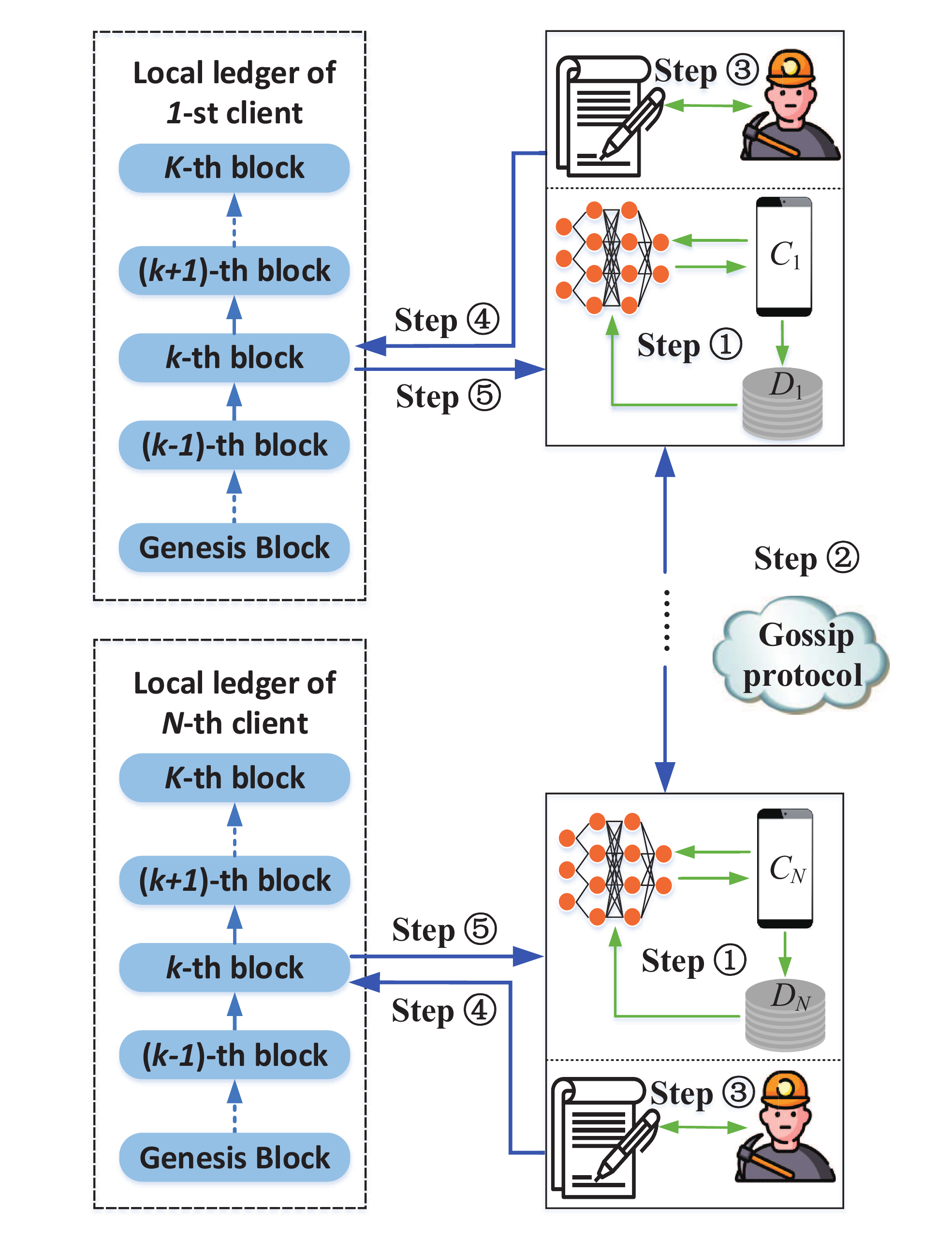}\\
  %\vspace{-0.3cm}
  \caption{Key steps in the $k$-th integrated round of the proposed BLADE-FL.
  }
  \vspace{-0.4cm}
  \label{graph_framework}
\end{figure}

Let us consider that all the clients deploy the same time allocation strategy for local training and mining.
In other words, all the clients start the training at the same time, and then turn to the mining stage simultaneously.
In this context, for each global model update and block generation, we define an \emph{integrated round} for BLADE-FL that combines a communication round of FL and a mining round of blockchain.
%First of all, each client initializes its local parameters, such as initial weight, learning rate, etc.. After the initialization, all the clients perform the procedure of training and mining.
As illustrated in Fig.~\ref{graph_framework}, the $k$-th integrated round can be specified as the following steps\footnote{In the very beginning of first integrated round, each client initializes its local parameters, such as initial weight, learning rate, etc..}.
%\vspace{-0.2cm}

\begin{basedescript}{\desclabelstyle{\pushlabel}\desclabelwidth{4em}}
\item[$\bullet$ \textbf{Step \large{\ding{172}}}:]
\emph{Local Training.} Each client performs the local training by iterating the learning algorithm $\tau$ times to update its own model $\boldsymbol{w}_i^k, \forall i$.
\item[$\bullet$ \textbf{Step \large{\ding{173}}}:]
\emph{Model Broadcasting and Verification.} Each client signs its models by the digital signature and propagates the models as its requested transactions. The other clients verify the transactions of the requested client (i.e., identity of the client).
\item[$\bullet$ \textbf{Step \large{\ding{174}}}:]
\emph{Mining.} Upon receiving the models from others, all the clients generate the global model and then compete to mine the $k$-th block.
\item[$\bullet$ \textbf{Step \large{\ding{175}}}:]
\emph{Block Validation.} All the clients append the new block onto their local ledgers only if the block is validated.
\item[$\bullet$ \textbf{Step \large{\ding{176}}}:]
\emph{Local Updating.} Upon receipt of verified transactions in this block, each client updates its local model. Then the system proceeds to the $(k+1)$-th round.
%Thus, the global aggregation is processed locally when a new block is generated. Then the system will turn to the next round.
\end{basedescript}

In contrast to~\cite{DBLP:journals/tii/LuHDMZ20a}, BLADE-FL does not rely on an additional third-party for global aggregation, thereby promoting privacy against model leakage. From the above steps, the consensus mechanism builds a bridge between the local models from clients and the model aggregation.
Thanks to PoW, BLADE-FL guarantees the tamper-resistant model update in a trusted blockchain network.

\subsection{Computing Resource Allocation Model}\label{subsec_CRA}
%The consensus mechanism builds a bridge between the local models from clients and the global model for performing model aggregation.
In this subsection, we model the time required for training and mining, to show the relationship between FL and blockchain in BLADE-FL.

\emph{Block Generation Rate:} The block generation rate is determined by the computation complexity of the hash function and the total computing power of the blockchain network~(i.e., total CPU cycles).
The average CPU cycles required to generate a block in PoW is defined as $\mathbb{E}[\mathrm{PoW}]=\kappa\chi$, where $\kappa$ is the mining difficulty\footnote{Following PoW, the mining difficulty is adjusted at different intervals but maintains unaltered over each interval. Thus, we consider that the average CPU is invariant over the period with a fixed mining difficulty.}, and $\chi$ denotes the average number of total CPU cycles to generate a block~\cite{DBLP:journals/tpds/XuWLGLYG19}. Thus, we define the average generation time of a block as
\begin{equation}\label{eq_Etbl}
\begin{aligned}
\beta \triangleq \frac{\mathbb{E}[\mathrm{PoW}]}{N f}
=\frac{\kappa\chi}{N f},
\end{aligned}
\end{equation}
where $f$ denotes the CPU cycles per second of each client. Given a fixed $f$, $\beta$ is a constant.

\emph{Local Training Rate:} Recall that the local training of each client contains $\tau$ iterations. The training time consumed by each training iteration at the $i$-th client is given by~\cite{9242286}
\begin{equation}\label{eq_Etloc}
\alpha_i \triangleq \frac{\vert \mathcal{D}_i\vert \rho}{f},
\end{equation}
where $\vert \mathcal{D}_i\vert$ denotes the number of samples in the $i$-th client, and $\rho$ denotes the number of CPU cycles required to train one sample.
This paper considers that each client is equipped with the same hardware resources (e.g., CPU, battery, and cache memory), and has sufficient energy. The performance analysis and optimization under limited energy will be considered in the future direction. Therefore, each client is loaded  with the same number of local samples, and has the same $f$ and $\rho$.  However, the contents of samples owned by different clients are diverse. For simplicity, we assume that each client uses the same training algorithm and trains the same number of $\tau$ iterations for its local model update.
Consequently, each client has an identical local training time per iteration.
In this context, we let $\alpha=\alpha_i, \forall i$, as a constant.

Consider that a typical FL learning task is required to be accomplished within a fixed duration of $t^{\mathrm{sum}}$. Given the same hardware configuration, each client has the total number of CPU cycles $ft^{\mathrm{sum}}$.
%Under this computing time constraint, it is crucial for each client to optimally allocate the total CPU cycles for both local training and mining.
From to (\ref{eq_Etbl}) and (\ref{eq_Etloc}), the number of iterations for local training in each integrated round is given by
\begin{equation}\label{eq_t_allocation}
\tau = \left\lfloor \frac{1}{\alpha} \left(\frac{t^{\mathrm{sum}}}{K }-\beta \right) \right\rfloor,
\end{equation}
where $\lfloor \cdot \rfloor$ denotes the floor function, and $K$ is a positive integer that represents the number of total integrated round.
Furthermore, $K\tau \alpha$ denotes the total training time, while $K \beta$ is the total mining time.
Under the constraint of computing time $ t^{\mathrm{sum}}$, we notice that the longer the mining takes, the shorter the training occupies. That is because that (\ref{eq_t_allocation}) implies a fundamental tradeoff in BLADE-FL, i.e., the more iterations each client trains locally, the fewer integrated rounds the BLADE-FL network performs.
Moreover, due to the floor operation in (\ref{eq_t_allocation}), there may exists some computing time left, i.e., $t^{\mathrm{sum}}-K(\tau \alpha+ \beta)\geq 0$. We stress that the extra time is not sufficient to perform another integrated round, and thereby the global model cannot update during this period. In this context, we ignore this computing time and assume $t^{\mathrm{sum}}-K(\tau \alpha+ \beta)= 0$ in the following analysis.

In what follows, we optimize the learning performance of BLADE-FL based on (\ref{eq_t_allocation}).

\section{Performance Analysis of the BLADE-FL System}\label{sec:Con_FL}

In this section, we evaluate the learning performance of BLADE-FL with the upper bound on the loss function in Section \ref{subsec_upperbound}, and optimize the learning performance with respect to the number of integrated rounds in Section \ref{subsec_optiaml_K}.

\subsection{Achievable Upper Bound Analysis}\label{subsec_upperbound}
Existing works such as [3]-[11] evaluated the learning performance of the standard FL based on the loss function, where a smaller value of the loss function corresponds to a learning model with higher accuracy. Recently,
the work in~\cite{DBLP:journals/jsac/WangTSLMHC19} derived an upper bound on the loss function between the iterations of local training and global aggregation.

Compared with the standard FL, our BLADE-FL replaces the centralized server with a blockchain network for global aggregation. Notably, the training process and the aggregation rule are the same as the centralized FL. Thus, the derived upper bound on the loss function in \cite{DBLP:journals/jsac/WangTSLMHC19} can be applied to BLADE-FL.

We make the following assumption for all the clients.
\begin{assumption}\label{assumption_1} %\label{assu:training calculation}
\begin{enumerate}[For any two different $\boldsymbol{w}$ and $\boldsymbol{w}'$, we assume that]
\item $F_i(\boldsymbol{w})$ is convex \cite{DBLP:journals/corr/abs-2010-05958}, i.e., $F(\boldsymbol{w})\geq F(\boldsymbol{w}')+\nabla F(\boldsymbol{w}')^T (\boldsymbol{w}-\boldsymbol{w}')$;
\item $F_i(\boldsymbol{w})$ is $\xi$-Lipschitz \cite{DBLP:conf/pkdd/KarimiNS16}, i.e., $\Vert F_i(\boldsymbol{w})-F_i(\boldsymbol{w}')\Vert_2\leq \xi \Vert\boldsymbol{w}-\boldsymbol{w}'\Vert_2$;
\item $F_i(\boldsymbol{w})$ is L-smooth \cite{DBLP:journals/corr/abs-2010-05958}, i.e., $\Vert\nabla F_i(\boldsymbol{w})-\nabla F_i(\boldsymbol{w}')\Vert_2\leq L \Vert\boldsymbol{w}-\boldsymbol{w}'\Vert_2$.
\end{enumerate}
\end{assumption}

According to Assumption~\ref{assumption_1}, $F(\boldsymbol{w})$ is \textit{convex, $\xi$-Lipschitz,} and \textit{L-smooth}~\cite{DBLP:conf/pkdd/KarimiNS16}.

The work in \cite{DBLP:journals/jsac/WangTSLMHC19} also defined the following definition of measurements to capture the divergence between the gradient of the local loss function and that of the global loss function.
\begin{definition}[(Gradient Divergence) \cite{DBLP:journals/jsac/WangTSLMHC19}]\label{definition_1}
For each client, we define $\delta_i$ as an upper bound on $\Vert\nabla F_i(\bar{\boldsymbol{w}})-\nabla F(\bar{\boldsymbol{w}})\Vert_2$, i.e., $\Vert\nabla F_i(\bar{\boldsymbol{w}})-\nabla F(\bar{\boldsymbol{w}})\Vert_2\leq \delta_i$. Thus, the global gradient divergence $\delta$ can be expressed as $\delta=\frac{\sum_i \vert D_i \vert \delta_i}{N}$.
\end{definition}

This divergence is related to the distribution of local datasets over different clients.

Based on the definition and assumptions, we derive the following theorem.
\begin{theorem}{The upper bound of loss function of BLADE-FL.}\label{theorem_upper}
\begin{equation}\label{final_equation}
\begin{split}
F(\bar{\boldsymbol{w}}^{K})-F(\bar{\boldsymbol{w}}^*)&\leq G(K, \alpha, \beta, \eta, \delta, t^{\mathrm{sum}})\\
&=\frac{1}{\gamma \left(\eta\phi-\frac{\frac{\delta\xi K}{L}\left( \lambda^{\frac{\gamma}{K}}-1\right)-\eta\xi\delta \gamma}{\varepsilon^2 \gamma}\right)},
\end{split}
\end{equation}
where
\begin{equation}
\lambda=\eta L+1, \quad \gamma=\frac{t^{\mathrm{sum}}-K\beta}{\alpha},
\end{equation}
and $G(K, \alpha, \beta, \eta, \delta, t^{\mathrm{sum}})$ denotes the upper bound on the loss function in BLADE-FL.
\end{theorem}
\begin{IEEEproof}
Please see Appendix \ref{appendix_theo1}.
\end{IEEEproof}

The upper bound in (\ref{final_equation}) shows that the learning performance depends on the total number of integrated rounds $K$, the local training time per iteration $\alpha$, the average mining time per block $\beta$, the learning rate $\eta$, the data distribution $\delta$, and the total computing time $t^{\mathrm{sum}}$.
From Definition \ref{definition_1}, $\delta$ is fixed given $N$ and the datasets of each client, and $\eta$ is preset.
Recall that $\alpha$ and $\beta$ are both constant in (\ref{eq_Etbl}) and (\ref{eq_Etloc}). Given any fixed $\delta$, $\eta$, $\alpha$, $\beta$ and $t^{\mathrm{sum}}$, $G(\cdot)$ in (\ref{final_equation}) is an univariate function of $K$.
In the following theorem, we verify that $G(K)$ is a convex function with respect to $K$.

\begin{theorem}\label{theorem_convex}
$G(K)$ is convex with respect to $K$.
\end{theorem}
\begin{IEEEproof}
Please see Appendix \ref{appendix_theo2}.
\end{IEEEproof}

Note that $K$ should not be too small, since a tiny $K$ will make the system vulnerable to external attacks \cite{9119406}.

\subsection{Optimal Computing Resource Allocation}\label{subsec_optiaml_K}
%In this subsection, we optimize the computing resource allocation method to minimize $G(K)$ when $\alpha$, $\beta$, $N$ (or $\delta)$ and $\eta$ are all fixed.

First, the following theorem shows the optimal solution that minimizes $G(K)$.

\begin{theorem}\label{theorem_closed}
Given any fixed $\alpha$, $\beta$, $N$ (or $\delta)$ and $\eta$, the optimal number of integrated rounds that minimizes the upper bound on the loss function in (\ref{final_equation}) is given by
\begin{equation}\label{closed_solution}
K^* = \frac{t^{\mathrm{sum}}}{\sqrt{\frac{2\alpha\beta}{\eta L}+ \alpha \beta+\beta^2}},
 \end{equation}
when $ \frac{\eta L\gamma}{K}\ll 1$.% and $\varepsilon^2\geq\frac{\delta\xi}{\phi}$.
\end{theorem}
\begin{IEEEproof}
Please see Appendix \ref{appendix_theo3}.
\end{IEEEproof}

Then, under a fixed constraint $t^{\mathrm{sum}}$, let us focus on the effect of $\alpha$ and $\beta$ on $K^*$ under fixed $N$ and $\eta$ by the following corollary\footnote{The following analytical results in Corollary 1, 2, 3, 4, and 5 are with respect to $K$. Due to the fundamental tradeoff between $K$ and $\tau$, the opposite results with respect to $\tau$ also hold. }.

\begin{corollary}\label{pro_difficulty}
Given $N$ and $\eta$, the optimal value $K$ decreases as either $\alpha$ or $\beta$ goes up. In this case, more time is allocated to training when $\alpha$ gets larger or to mining when $\beta$ becomes larger.
\end{corollary}
\begin{IEEEproof}
This corollary is a straightforward result from Theorem \ref{theorem_closed}.
\end{IEEEproof}

Recall that $\alpha$ denotes the training time per iteration, and $\beta$ denotes the mining time per block. From Corollary \ref{pro_difficulty}, the longer a local training iteration takes, the more computing power allocated to the local training at each client. Similarly, each client allocates more computing power to the mining when the mining time is larger.

Next, we investigate the impact of $N$ and $\eta$ on $K^*$ when $\alpha$ and $\beta$ are fixed by the following corollaries (i.e., Corollary \ref{delta_remark} and Corollary \ref{eta_remark}).

\begin{corollary}\label{delta_remark}
Given fixed $\alpha$ and $\beta$, $K^*$ becomes larger as $\delta$ grows. In this case, more time is allocated to the mining.
\end{corollary}
\begin{IEEEproof}
Please see Appendix \ref{appendix_corodelta}.
\end{IEEEproof}

\begin{corollary}\label{proposition_1}
Given fixed $\alpha$ and $\beta$, $K^*$ becomes smaller as $N$ grows. In this case, more time is allocated to the training.
\end{corollary}

\begin{IEEEproof}
Based on Corollary \ref{delta_remark}, the proof of Corollary \ref{proposition_1} is straightforward, since $\delta$ drops as $N$ grows.
\end{IEEEproof}

The explanation of Corollary \ref{proposition_1} is that each client may have trained an accurate local model but not an accurate global model~($\delta$ is large), and thus BLADE-FL needs to perform more global aggregation especially when $N$ is small. This paper considers a number of honest clients in BLADE-FL to defend the malicious mining \cite{DBLP:conf/trustbus/AbramsonHPPB20}.
When $N$ is sufficiently large, $\delta$ converges to its mean value according to the law of large number. In this context, Corollary \ref{delta_remark}  shows that $K^*$ approaches a constant as $\delta$ converges, and further implies that $K^*$ is independent of $N$.

%Afterwards, we investigate the impact of $\eta$ on $K^*$ when $\alpha$, $\beta$, and $N$ are fixed by the following corollary.
\begin{corollary}\label{eta_remark}
Given fixed $\alpha$ and $\beta$, $K^*$ increases as $\eta$ goes larger. Meanwhile, the upper bound in (\ref{final_equation}) drops as $\eta$ grows if $\eta L<1$.
\end{corollary}
\begin{IEEEproof}
Please see Appendix \ref{appendix_coroeta}.
\end{IEEEproof}

The reason behind Corollary \ref{eta_remark} is that the global model may not converge when each client is allocated with limited learning resources and a small learning rate.
In addition, a higher learning rate may lead to faster convergence but a less inaccurate local model.
To compensate for the inaccurate training, more computing power is allocated to the local training.
In practice, the learning rate is decided by the learning algorithm, and the learning rates of different learning algorithms are diverse. Therefore, we can treat $\eta$ as a constant in BLADE-FL.

\section{Performance Analysis with Lazy clients}\label{Sec:Lazy}
Different from the conventional FL, a new problem of learning deficiency caused by lazy clients emerges in the BLADE-FL system. This issue is fundamentally originated from the lack of an effective detection and penalty mechanism in an unsupervised network such as blockchain, where the lazy client is able to plagiarize models from others to save its own computing power.
The lazy client does not contribute to the global aggregation, and even causes training deficiency and performance degradation.
To study this issue, we first model the lazy client in Section \ref{subsec_lazymodel}. Then, we develop an upper bound on the loss function to evaluate the learning performance of BLADE-FL with the presence of lazy clients in Section \ref{subsec_lazy_upperbound}. Next, we investigate the impact of the ratio of lazy clients and the power
of artificial noises on the learning performance in Section \ref{subsec_lazy_performance}. In this section, suppose that there exist $M$ lazy clients in BLADE-FL and $M \leq N$. Let us define the lazy ratio as $\frac{M}{N}$.

\subsection{Model of Lazy Clients}\label{subsec_lazymodel}
\begin{figure}[t]%%%%%%%%%%%%%%%%%½ûֹͼƬ¸¡¶¯
  \centering
  % Requires \usepackage{float}À´½ûÖ¹¸¡¶¯
  \includegraphics[width=0.5\textwidth]{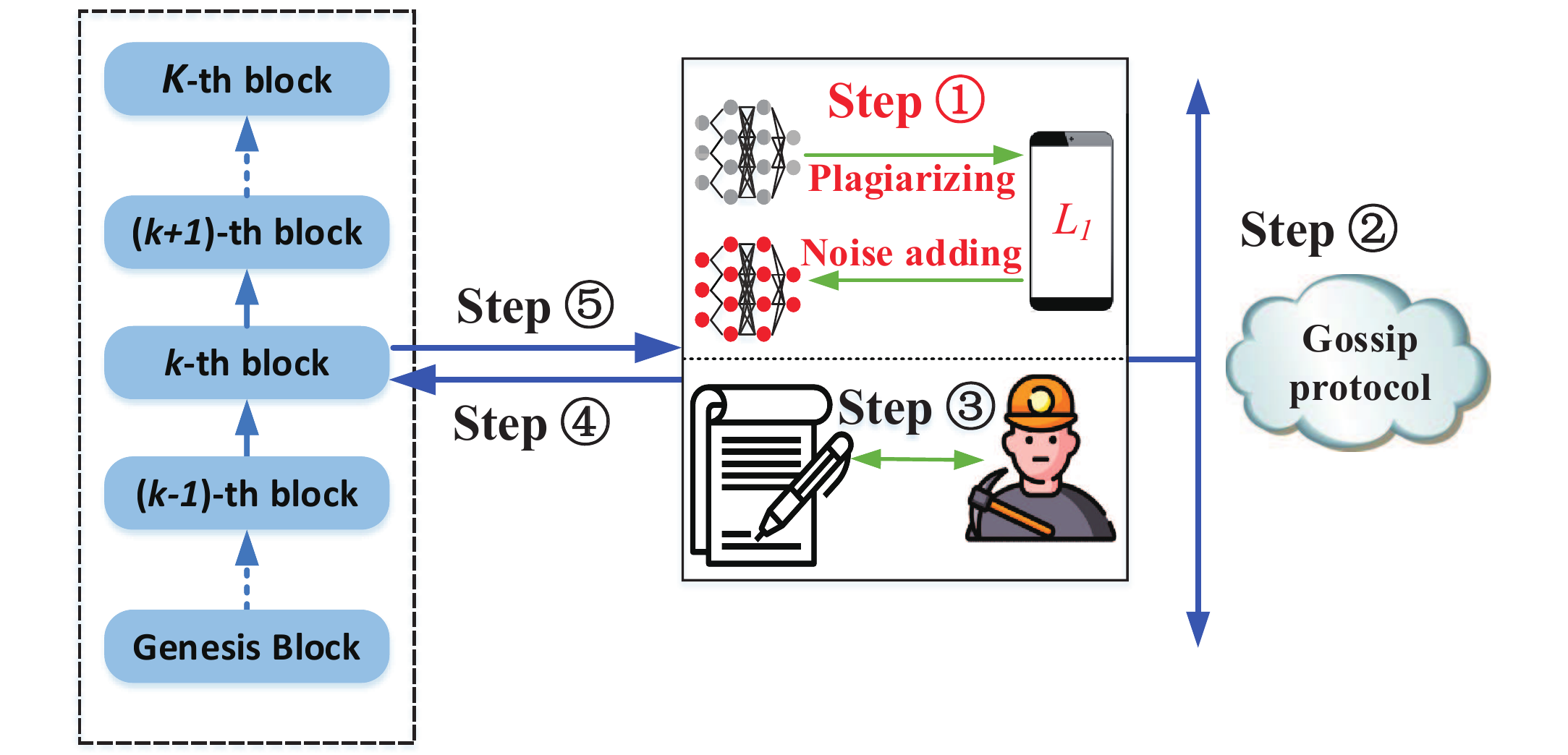}\\
  %\vspace{-0.1cm}
  \caption{Model of the lazy client in BLADE-FL. %Instead of local training, lazy clients will plagiarize others' trained model and add artificial noise onto it.
  }
  %\vspace{-0.3cm}
    \label{fig_lazy}
\end{figure}

A lazy client can simply plagiarize other models before mining a new block. To avoid being spotted by the system, each lazy client adds artificial noises to its model weights as
\begin{equation}\label{lazy_client}
\boldsymbol{w}_{i'}^{k}=\boldsymbol{w}_i^{k}+\mathbf{n}_i,\quad i'\subseteq \mathcal{M}, i\nsubseteq \mathcal{M}, k=1,2,\dots,K,
\end{equation}
where $\mathcal{M}$ denotes the set of lazy clients, $\mathbf{n}_i$ is the artificial noise vector following a Gaussian distribution with mean zero and variance $\sigma^2$. As Fig.~\ref{fig_lazy} illustrates, the $i$-th client is identified as the lazy client if it plagiarizes an uploaded model from others and add artificial noise onto it in \textbf{Step} {\large{\ding{172}}}.
Except the plagiarism in \textbf{Step} {\large{\ding{172}}}, the lazy clients follow the honest clients %\footnote{In this paper, we assume that each client is honest in the mining, because of the mining reward. }
to perform \textbf{Step} {\large{\ding{173}-\ding{176}}}.

%\vspace{-0.25cm}

 \subsection{Achievable Upper Bound with Lazy Clients}\label{subsec_lazy_upperbound}
In this subsection, we develop an upper bound on the loss function with the lazy ratio and the power of artificial noise in the following theorem.

\begin{theorem}\label{theorem_lazyupper}
Using the model of lazy clients in (\ref{lazy_client}), an upper bound on the loss function after $K$ integrated rounds with the lazy ratio of $\frac{M}{N}$ is given by

\begin{equation}\label{eq_lazy_sigma}
\begin{split}
&F(\tilde{\boldsymbol{w}}^{K})-F(\bar{\boldsymbol{w}}^*) \leq \tilde{G}(K, \alpha, \beta, \eta, \delta, t^{\mathrm{sum}}, \theta, \sigma^2)\\
&= \frac{1}{\gamma \left(\eta\phi-\frac{\frac{\delta\xi K}{L}\left( \lambda^{\frac{\gamma}{K}}-1\right)-\eta\xi\delta \gamma+K \xi \frac{M}{N}\theta+K\xi \frac{\sqrt{M}}{N}\sigma^2}{\varepsilon^2 \gamma}\right)},
\end{split}
\end{equation}
\end{theorem}
where $\tilde{\boldsymbol{w}}^{K}$ denotes the aggregated weights of BLADE-FL with lazy clients after $K$ integrated rounds, and
$\theta = \Vert\boldsymbol{w}_{i'}^{K}-\tilde{\boldsymbol{w}}_{i'}^{K}\Vert_2$
denotes the performance degradation caused by lazy clients after $K$ integrated rounds.
\begin{IEEEproof}
Please see Appendix \ref{appendix_theolazy}.
\end{IEEEproof}

Thereafter, we use the upper bound in~(\ref{eq_lazy_sigma}) to evaluate the learning performance of BLADE-FL with lazy clients.

\subsection{Optimization on Performance with Lazy Clients}\label{subsec_lazy_performance}
Practically, a lazy node tends not to add either huge or tiny noise in order to conceal itself. To this end, it is required that the value of $\sigma^2$ is comparable to that of $\theta$.

\begin{remark}
From (\ref{eq_lazy_sigma}), the plagiarism behavior contributes a term proportional to $\frac{M}{N}$ to the bound, while the artificial noise exhibits an impact term proportional to $\frac{\sqrt{M}}{N}$. This
%is because the plagiarism behavior is related to the original data while the artificial noise is independent of the original data, which
indicates that the plagiarism has a more significant effect on the learning performance compared with the noise perturbation.
\end{remark}

Then, we analyze the impact of $\frac{M}{N}$ and $\sigma^2$ on the optimal value of $K$ in the following corollary.
\begin{corollary}\label{pro_lazy_sigma}
The optimal $K$ that minimizes $\tilde{G}(\cdot)$ in (\ref{eq_lazy_sigma}) decreases as either the lazy ratio $\frac{M}{N}$ or the noise variance $\sigma^2$ grows.
\end{corollary}
%\begin{figure}[t]%%%%%%%%%%%%%%%%%½ûֹͼƬ¸¡¶¯
%  \centering
%  \includegraphics[width=0.4\textwidth, height=0.17\textwidth ]{Written_digits_modified.png}\\
%  \vspace{-0.1cm}
%  \caption{Visual illustration of the standard MNIST dataset}
%  \vspace{-0.3cm}
%    \label{fig_2}
%\end{figure}
\begin{IEEEproof}
Please see Appendix \ref{appendix_corolazy}.
\end{IEEEproof}

When the system is infested with a large number of lazy clients (i.e., the lazy ratio $\frac{M}{N}$ approaches 1), more computing power should be allocated to local training to compensate for the insufficient learning.

%\begin{figure}[t]%%%%%%%%%%%%%%%%%½ûֹͼƬ¸¡¶¯
%  \centering
%  \includegraphics[width=0.4\textwidth, height=0.17\textwidth]{Fashion_MNIST_modified.png}\\
%  \vspace{-0.1cm}
%  \caption{Visual illustration of the Fashion-MNIST dataset.}
%  \vspace{-0.1cm}
%    \label{fig_Fashion}
%\end{figure}

\section{Discussions on Privacy Issue}\label{Sec:privacy}
In BLADE-FL, each client can add noise into local model as differential privacy~(DP) mechanism \cite{DBLP:journals/tifs/WeiLDMYFJQP20} to meet the need of privacy. For example, $\epsilon$-DP mechanism provides a strong criterion for the privacy preservation schemes.
Here, $\epsilon > 0$ is the distinguishable bound of all outputs on neighboring datasets $\mathcal D, \mathcal D'$ in a database. A larger $\epsilon$ gives a clearer distinguishability of neighboring datasets and thereby a higher risk of privacy violation.
The definition of $\epsilon$-DP mechanism is as follows.
\begin{definition}($\epsilon$-DP~\cite{DBLP:journals/iotj/ChamikaraBKLCA20}):
A randomized mechanism $\mathcal M: \mathcal{X}\rightarrow \mathcal{R}$ with domain $\mathcal{X}$ and range $\mathcal{R}$ satisfies $\epsilon$-DP,
if for all measurable sets $\mathcal S\subseteq \mathcal{R}$ and for any two adjacent datasets $\mathcal D, \mathcal D'\in \mathcal{X}$,
\begin{equation}
\emph{Pr}[\mathcal M(\mathcal D)\in \mathcal S]\leq e^{\epsilon}\emph{Pr}[\mathcal M(\mathcal D')\in \mathcal S].
\end{equation}
\end{definition}

The inner privacy issue can be referred to privacy preserving schemes for decentralized FL in other works~\cite{9055478, 8919319}.
When applying DP mechanism into BLADE-FL, we add a Gaussian noise with a small variance. Note that the DP mechanism solves the privacy leakage problem of FL, while our scheme solves the robust learning problem. These two problems are different and can be treated separately.
Besides, based on the developed upper bound in {Theorem~\ref{theorem_lazyupper}}, we derive that a small noise does not affect the optimal integrated round. The optimal computing power allocation stays the same whether add DP or not. Thus, Clients in BLADE-FL can use DP mechanism to promoting their privacy without affecting the optimal computing power allocation of the system.
We show the experimental results in Subsection \ref{appendix_expdp}.

\section{Experimental Results}\label{sec:Exm_Res}
In this section, we evaluate the analytical results with various learning parameters under limited computing time $t^{\mathrm{sum}}$. First, we evaluate the developed upper bound in (\ref{final_equation}), and then investigate the optimal value of overall integrated rounds $K$ under training time per iteration $\alpha$, mining time per block $\beta$, number of clients $N$, learning rate $\eta$, the ratio $\frac{M}{N}$, and power of artificial noise $\sigma^2$.

\subsection{Experimental Setting}
\begin{figure}[t]%%%%%%%%%%%%%%%%%½ûֹͼƬ¸¡¶¯
  \centering
  % Requires \usepackage{float}À´½ûÖ¹¸¡¶¯
  \includegraphics[height=0.3\textwidth, width=0.44\textwidth]{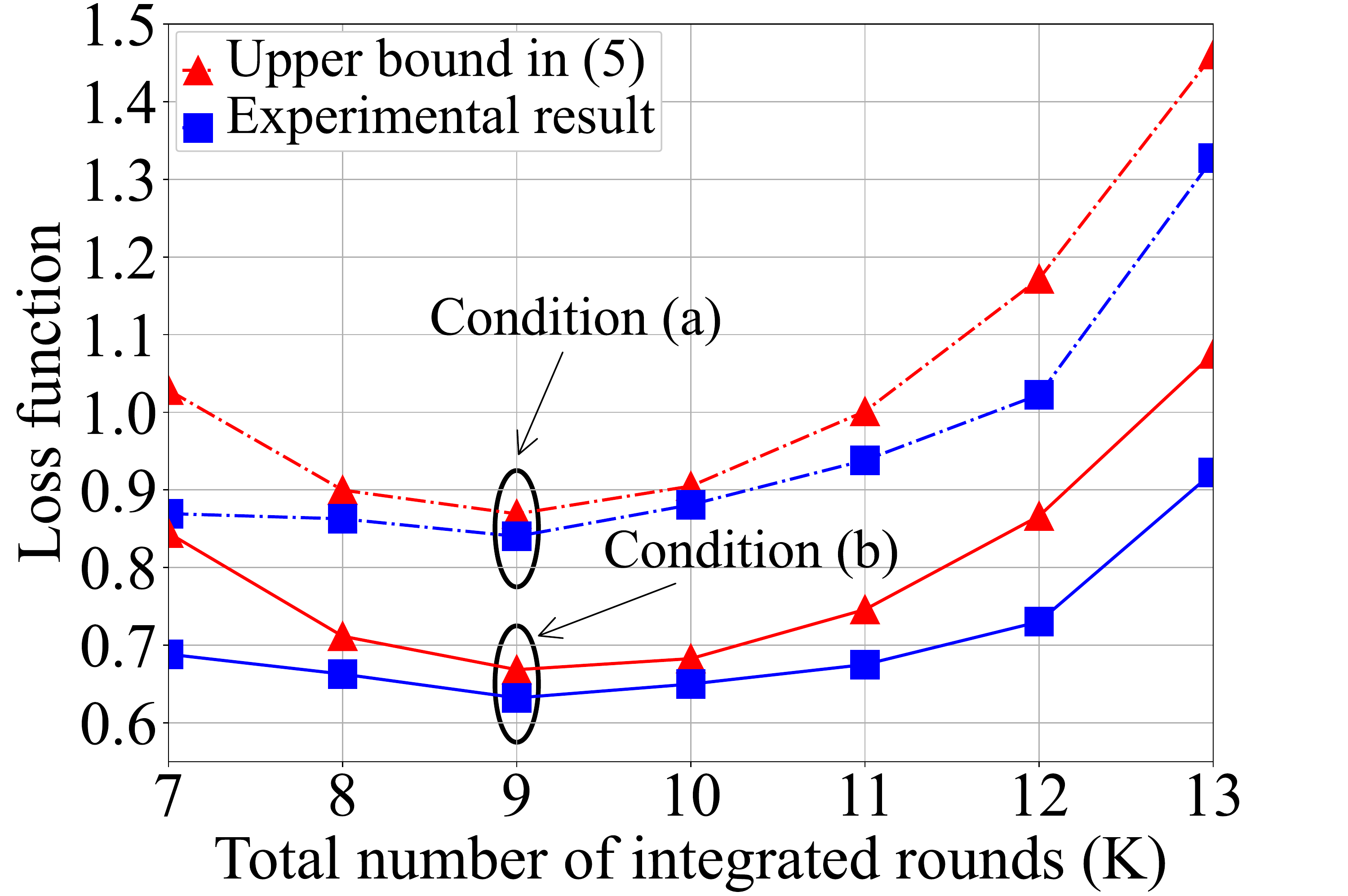}\\
  %\vspace{-0.1cm}
  \caption{The comparison of the upper bound in (\ref{final_equation}) and the experimental results for (a) $\alpha=1$, $\beta=6$, $\eta=0.005$, $N=20$, $M=0$ (b) $\alpha=1$, $\beta=6$, $\eta=0.010$, $N=20$, $M=8$.}
  %\vspace{-0.4cm}
    \label{fig_3}
\end{figure}

1) Datasets: In our experiments, we use two datasets for \textit{non-IID} setting to demonstrate the loss function and accuracy versus different values of $K$.

$\bullet$ MNIST. Standard MNIST handwritten digit recognition dataset consists of 60,000 training examples and 10,000 testing examples. %~\cite{726791}.
Each example is a 28$\times$28 sized handwritten digit in grayscale format from 0 to 9. %In Fig.~\ref{fig_2}, we illustrate several samples from the standard MNIST dataset.

$\bullet$ Fashion-MNIST. Fashion-MNIST for clothes has 10 different types, such as T-shirt, trousers, pullover, dress, coat, sandal, shirt, sneaker, bag, and ankle boot. %in Fig.~\ref{fig_Fashion}.

2) FL setting. Each client progresses the learning of a Multi-Layer Perceptron~(MLP) model. The MLP network has a single hidden layer that contains 256 hidden units. Each unit applies softmax function and rectified linear units of 10 classes~(corresponding to the 10 digits in MNIST and 10 clothes in Fashion-MNIST).

3) Parameters setting. In our experiments, we set the total computing time $t^{\mathrm{sum}}=100$, the samples of each client $\vert\mathcal{D}_i\vert=512,\forall i$, the number of clients $N=20$, the mining time per block $\beta=10$, the number of lazy clients $M=0$, and the learning rate $\eta=0.01$ as default, where the time is normalized by the training time per iteration $\alpha$.

\subsection{Experiments on Performance of BLADE-FL}

\begin{figure}[t]
  \centering
  \subfigure[]{
%  \begin{minipage}[t]{0.5\textwidth}
%  \centering
  \includegraphics[height=0.3\textwidth, width=0.44\textwidth]{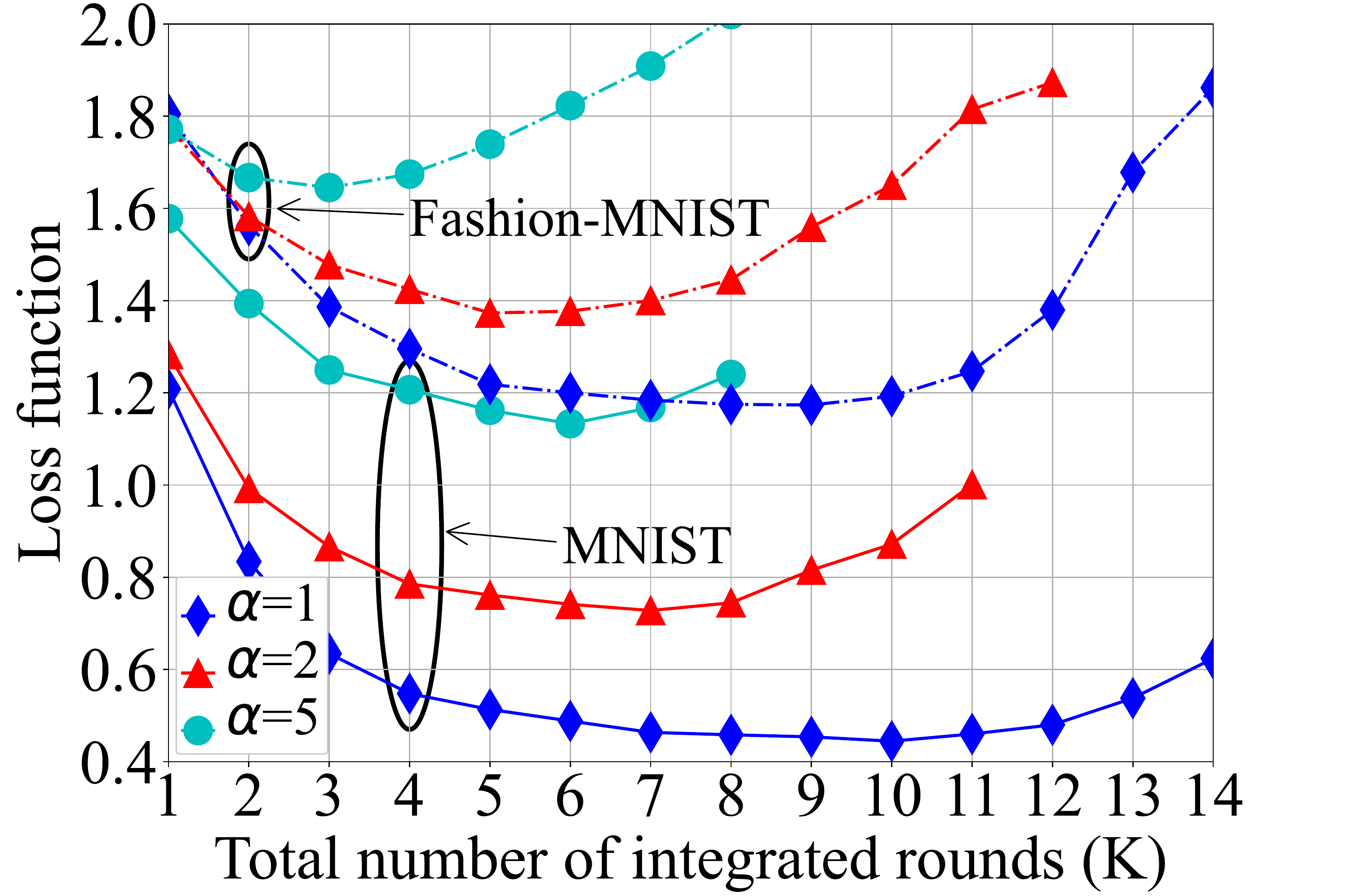}
%  \end{minipage}
  }
  \subfigure[]{
%  \begin{minipage}[t]{0.5\textwidth}
%  \centering
  \includegraphics[height=0.3\textwidth, width=0.44\textwidth]{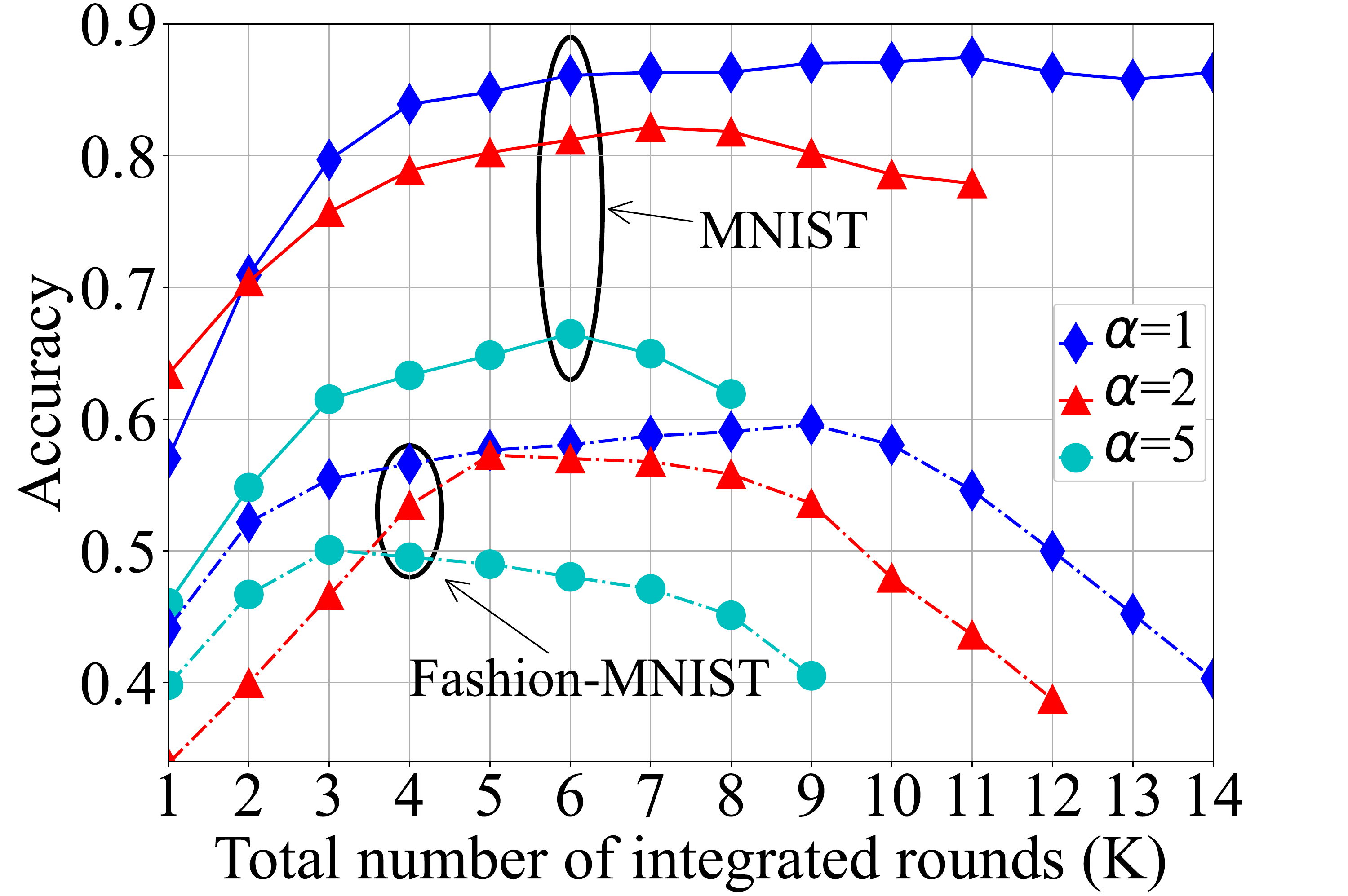}
%  \end{minipage}
  }
  \vspace{-0.3cm}
  \caption{Loss function and accuracy versus $K$ for different values of $\alpha$ for MNIST and Fashion-MNIST.}
  %\vspace{-0.2cm}
    \label{fig_FM_loss}
\end{figure}

\begin{table}[t]
\centering
\caption{The optimal training time and corresponding accuracy for different values of $\alpha$. }
\vspace{-0.1cm}
\begin{tabular}{c|c|c|c|c}
\hline
\multirow{3}{*}{\tabincell{c}{Training time  \\ per iteration}} &  \multicolumn{2}{c|}{$\tau \alpha K^*$}& \multicolumn{2}{c}{Maximal accuracy}\\
\cline{2-5}
&\multirow{2}{*}{MNIST} & Fashion- & \multirow{2}{*}{MNIST}& Fashion-\\
& & MNIST & & MNIST\\
\hline
$\alpha=1$ &40 & 46 & 87.44\% & 59.57\%  \\
\hline
$\alpha=2$ & 58 & 70 & 82.16\% & 57.18\% \\
\hline
${\alpha}=5$ & 64 & 82 & 66.47\% & 50.11\% \\
\hline
\end{tabular}
%\vspace{-0.3cm}
\label{tab:alpha}
\end{table}

\begin{figure}[t]
  \centering
  \subfigure[]{
%  \begin{minipage}[t]{0.5\textwidth}
%  \centering
  \includegraphics[height=0.3\textwidth, width=0.44\textwidth]{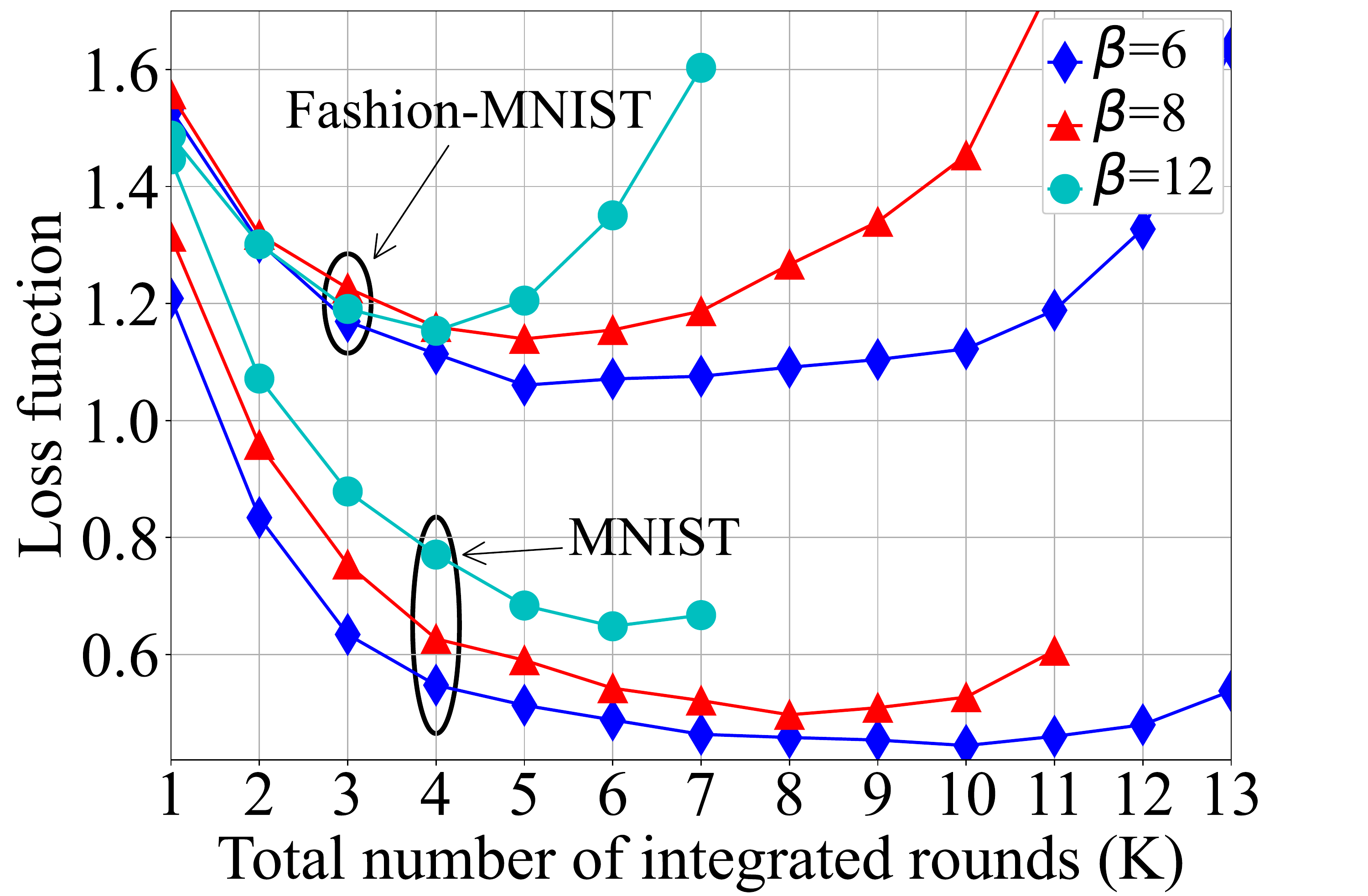}
%  \end{minipage}
  }
  %}
  \subfigure[]{
%  \begin{minipage}[t]{0.5\textwidth}
%  \centering
  \includegraphics[height=0.3\textwidth, width=0.44\textwidth]{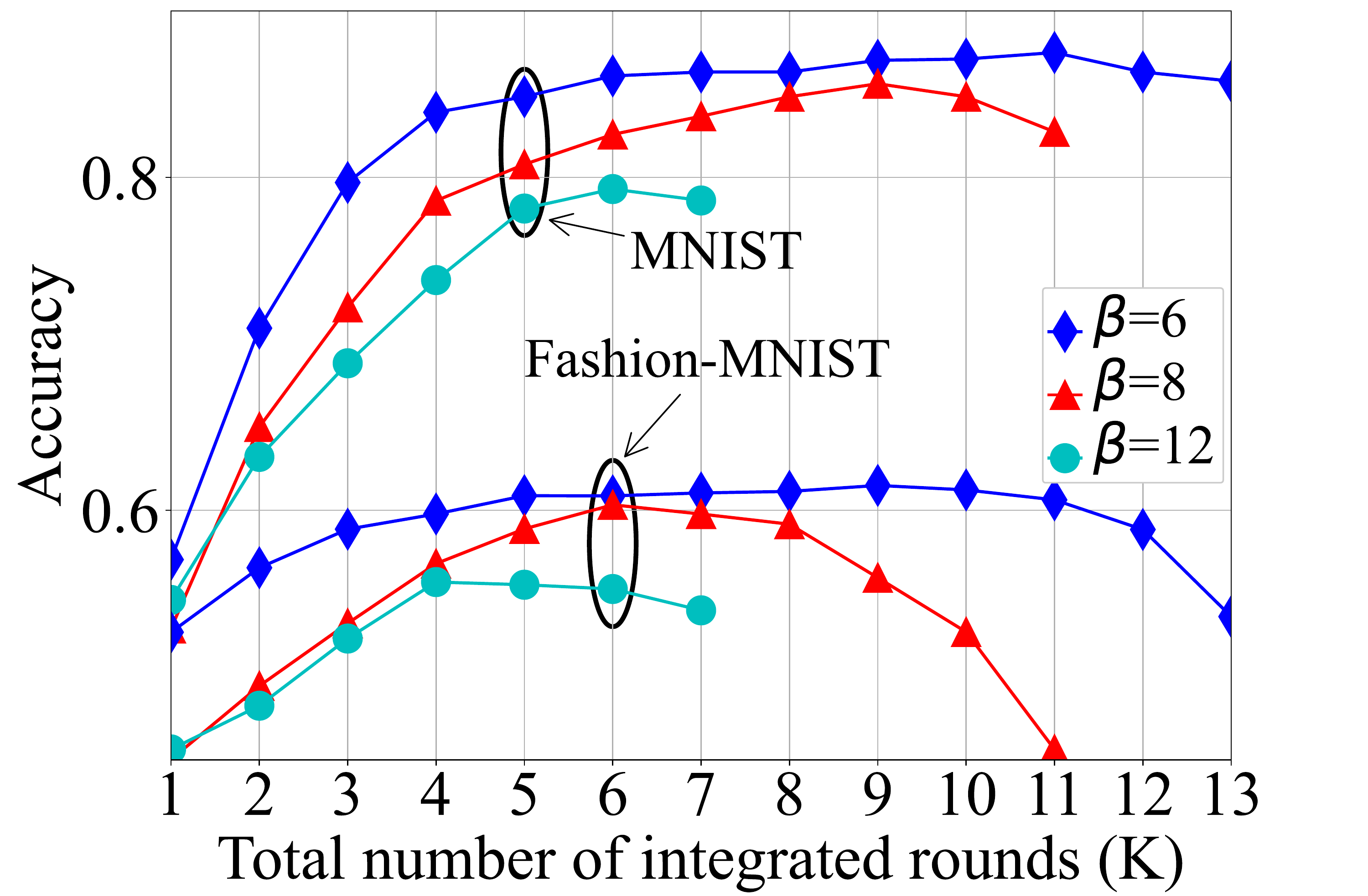}
%  \end{minipage}
  }
  \vspace{-0.3cm}
  \caption{Loss function and accuracy versus $K$ for different values of $\beta$ for MNIST and Fashion-MNIST.}
  \vspace{-0.15cm}
  \label{fig_7}
\end{figure}

\begin{table}[t]
\centering
\caption{The optimal mining time and corresponding accuracy for different values of $\beta$. }
\vspace{-0.1cm}
\begin{tabular}{c|c|c|c|c}
\hline
\multirow{3}{*}{\tabincell{c}{Mining time \\ per block}} &  \multicolumn{2}{c|}{$\beta K^*$}& \multicolumn{2}{c}{Maximal accuracy}\\
\cline{2-5}
&\multirow{2}{*}{MNIST} & Fashion- & \multirow{2}{*}{MNIST}& Fashion-\\
& & MNIST & & MNIST\\
\hline
$\beta=6$ &60 & 30 & 87.47\% & 61.51\%  \\
\hline
$\beta=8$ & 64 & 40 & 85.68\% & 60.34\% \\
\hline
$\beta=12$ & 72 & 48 & 79.32\% & 55.68\% \\
\hline
\end{tabular}
\vspace{-0.3cm}
\label{tab:beta}
\end{table}

Fig.~\ref{fig_3} plots the gap between the developed upper bound in (\ref{final_equation}) and the experimental results. We set learning rate $\eta=0.005$ and lazy ratio $\frac{M}{N}=0.4$ in conditions (a) and (b), respectively.
First, we can see that the developed bound is close but always higher than the experimental one under both conditions. Second, both the developed upper bound and the experimental results are convex with respect to $K$, which agrees with Theorem \ref{theorem_convex}. Third, both the upper bound in (\ref{final_equation}) and the experimental results reach the minimum at the same optimal value of $K$.

\begin{figure}[t]
  \centering
  \subfigure[]{
%  \begin{minipage}[t]{0.5\textwidth}
%  \centering
  \includegraphics[height=0.3\textwidth, width=0.44\textwidth]{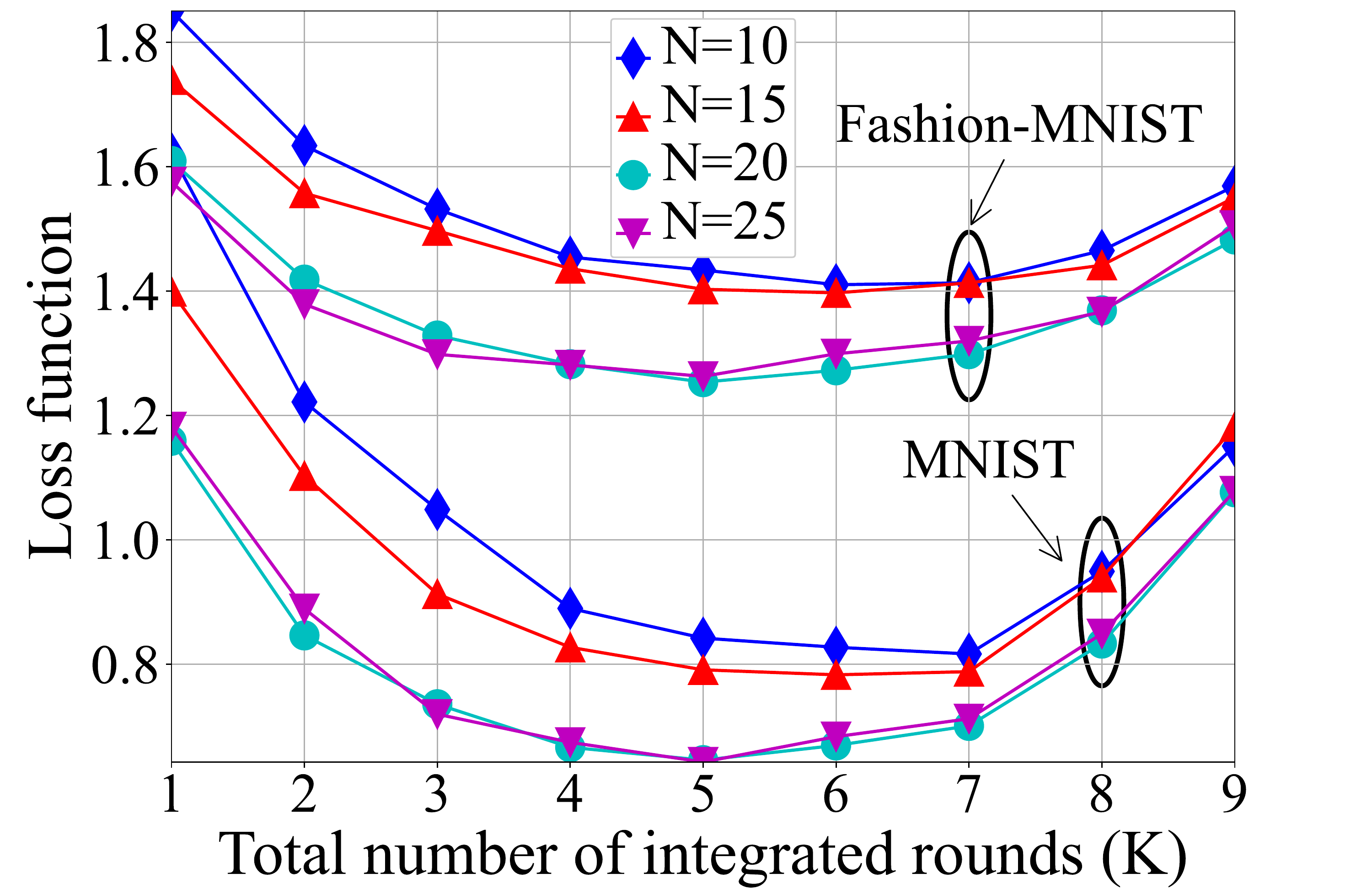}
%  \end{minipage}
  }
  %}
  \subfigure[]{
%  \begin{minipage}[t]{0.5\textwidth}
%  \centering
  \includegraphics[height=0.3\textwidth, width=0.44\textwidth]{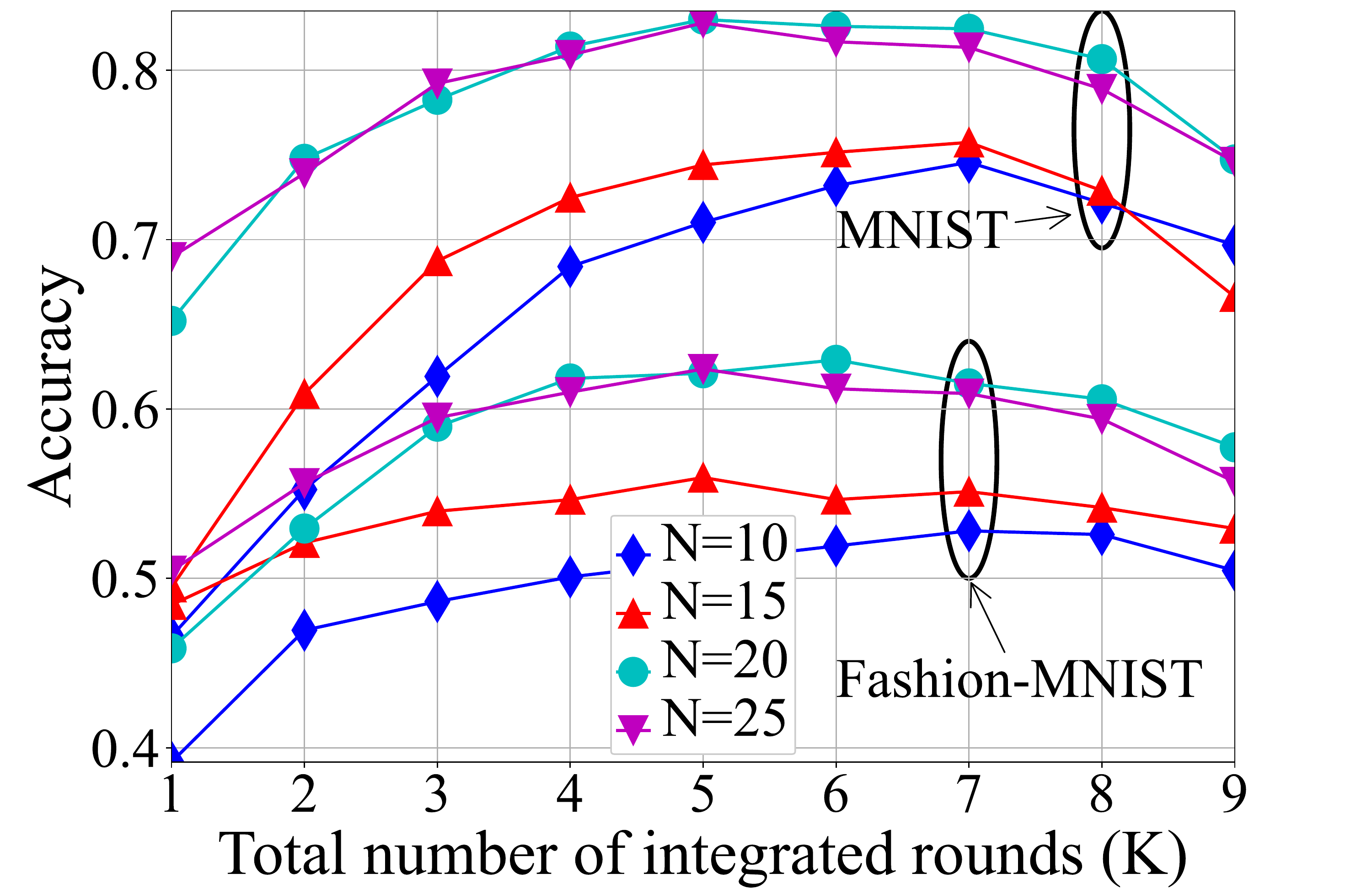}
%  \end{minipage}
  }
  \vspace{-0.3cm}
  \caption{Loss function and accuracy versus $K$ for different values of $N$ for MNIST and Fashion-MNIST.}
  %\vspace{-0.2cm}
    \label{fig_4}
\end{figure}

\begin{table}[t]
\centering
\caption{The optimal mining time and corresponding accuracy for different values of $N$.}
\vspace{-0.1cm}
\begin{tabular}{c|c|c|c|c}
\hline
\multirow{3}{*}{\tabincell{c}{Number of\\ clients}} &  \multicolumn{2}{c|}{$\beta K^*$}& \multicolumn{2}{c}{Maximal accuracy}\\
\cline{2-5}
&\multirow{2}{*}{MNIST} & Fashion- & \multirow{2}{*}{MNIST}& Fashion-\\
& & MNIST & & MNIST\\
\hline
N=10  & 70  &  42 & 74.52\% & 52.66\%\\
\hline
N=15  & 60 & 36 & 75.74\% & 55.83\%\\
\hline
N=20  & 50  &30 & 82.89\% & 62.91\%\\
\hline
N=25  & 50  &30 & 83.03\% & 62.64\%\\
\hline
\end{tabular}
%\vspace{-0.3cm}
\label{tab:N}
\end{table}

Fig.~\ref{fig_FM_loss} plots the experimental results of the loss function and accuracy on MNIST and Fashion-MNIST for different values of values of $\alpha$~(some curves are relatively short due to the limited total computing time), while Table~\ref{tab:alpha} shows the optimal training time and corresponding accuracy. Here, we set $\beta=6$. First, Fig.~\ref{fig_FM_loss}(a) shows that larger $\alpha$ leads to larger loss function. This is due to the fact that both $K$ and $\tau$ from (\ref{eq_t_allocation}) drops as $\alpha$ grows.
Second, from Table~\ref{tab:alpha}, the longer a training iteration consumes, the more training time each client takes. For example, using MNIST, the training time increases from $40$ to $64$ as $\alpha$ rises from $1$ to $5$. This observation is consistent with Corollary \ref{pro_difficulty}.

Fig.~\ref{fig_7} plots the experimental results of the loss function and accuracy on MNIST and Fashion-MNIST for different values of values of $\beta$~(some curves are relatively short due to the limited total computing time), while Table~\ref{tab:beta} shows the optimal mining time and corresponding accuracy.
First, Fig.~\ref{fig_7}(a) shows that larger $\beta$ leads to larger loss function, since both $K$ and $\tau$ from (\ref{eq_t_allocation}) drops as $\beta$ grows.
Second, from Table~\ref{tab:beta}, $K^*$ reduces as $\beta$ rises, but the optimal mining time $\beta K^*$ goes up as $\beta$ rises. For example, using MNIST, the mining time increases from $60$ to $72$ as $\beta$ grows from $6$ to $12$. This observation agrees with Corollary \ref{pro_difficulty}.

\begin{figure}[t]
  \centering
  \subfigure[]{
%  \begin{minipage}[t]{0.5\textwidth}
%  \centering
  \includegraphics[height=0.3\textwidth, width=0.44\textwidth]{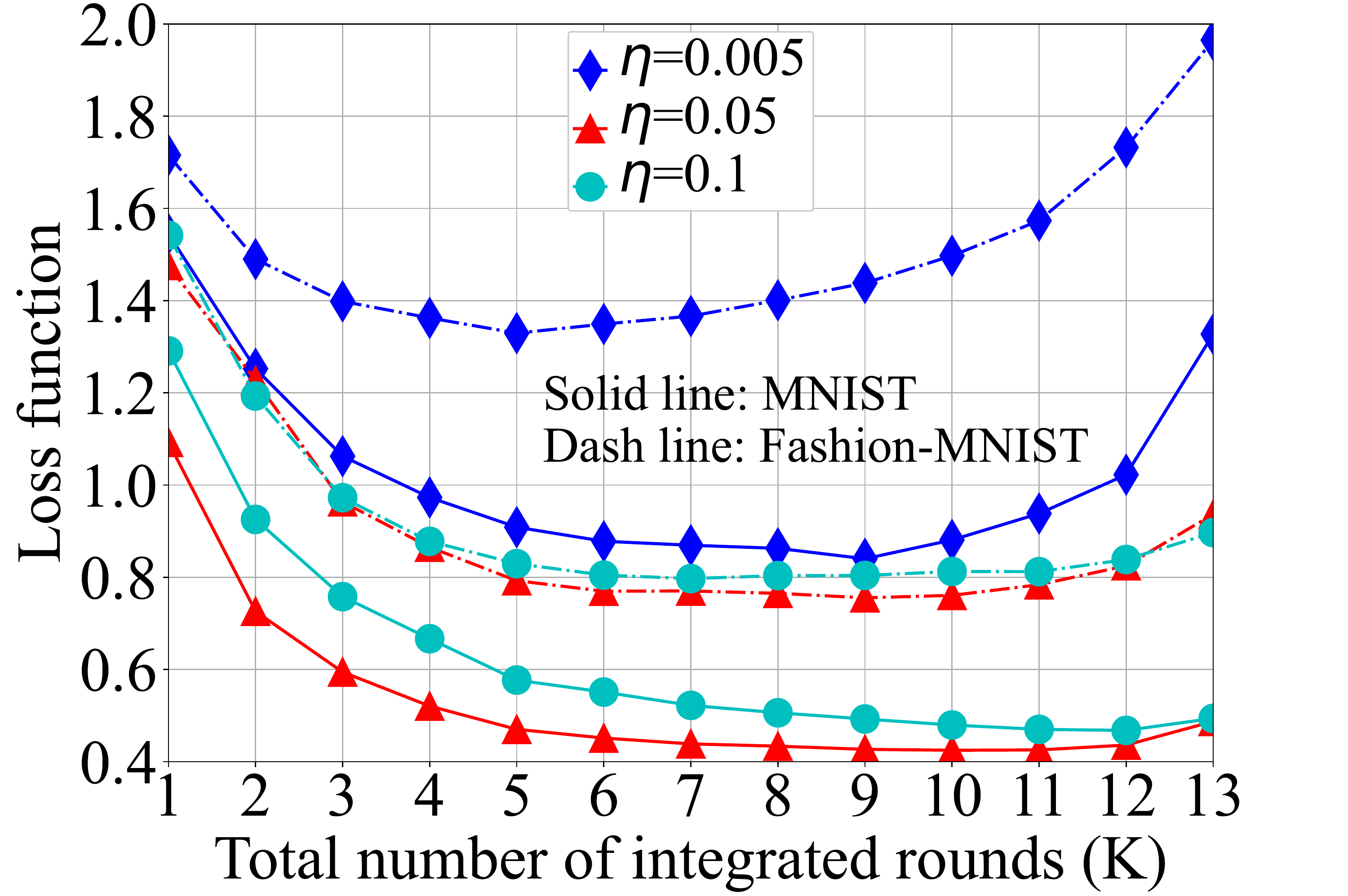}
%  \end{minipage}
  }
  %}
  \subfigure[]{
%  \begin{minipage}[t]{0.5\textwidth}
%  \centering
  \includegraphics[height=0.3\textwidth, width=0.44\textwidth]{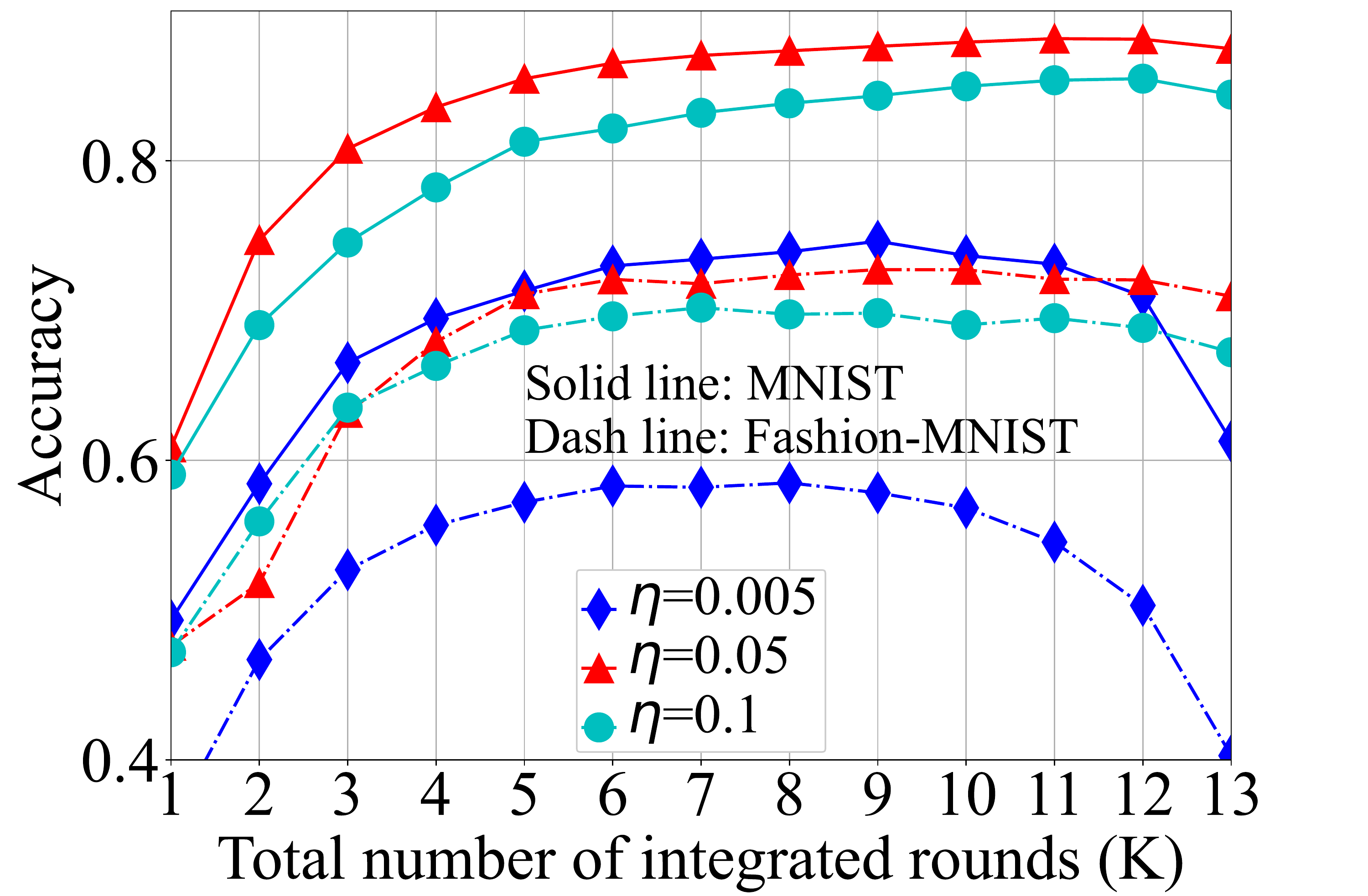}
%  \end{minipage}
  }
  %}
  \vspace{-0.3cm}
  \caption{Loss function and accuracy versus $K$ for different values of $\eta$ for MNIST and Fashion-MNIST. }
  %\vspace{0.2cm}
    \label{fig_5}
\end{figure}

\begin{table}[t]
\centering
\caption{The optimal mining time and corresponding accuracy for different values of $\eta$.}
\vspace{-0.1cm}
\begin{tabular}{c|c|c|c|c}
\hline
\multirow{3}{*}{\tabincell{c}{Learning\\ rate}} &  \multicolumn{2}{c|}{$\beta K^*$}& \multicolumn{2}{c}{Maximal accuracy}\\
\cline{2-5}
&\multirow{2}{*}{MNIST} & Fashion- & \multirow{2}{*}{MNIST}& Fashion-\\
& & MNIST & & MNIST\\
\hline
$\eta=0.005$  & 54  &  30& 74.70\% & 58.57\%\\
\hline
$\eta=0.05$  & 60 & 54& 88.17\% & 72.50\%\\
\hline
$\eta=0.1$  & 72  & 42& 85.51\% & 70.14\%\\
\hline
\end{tabular}
\vspace{-0.1cm}
\label{tab:eta}
\end{table}

Fig.~\ref{fig_4} shows the experimental results of the loss function and accuracy on MNIST and Fashion-MNIST for different values of values of $N$, while Table~\ref{tab:N} illustrates the optimal mining time and corresponding accuracy.
We set $\beta=6$. First, from Table~\ref{tab:N}, we notice that the optimal mining time drops as $N$ increases, which is consistent with Proposition \ref{proposition_1}. For example, using MNIST, $\beta K^*$ drops from $70$ to $50$ as $N$ rises from $10$ to $25$.
Second, from Fig.~\ref{fig_4}(a), larger $N$ leads to lower loss function. This is because the involved datasets are larger as $N$ grows, which causes a smaller loss function. Third, from both Fig.~\ref{fig_4}(a) and (b), $K^*$ approaches a fixed value when $N$ is sufficiently large (e.g., $N>20$). This observation is in line with Corollary \ref{proposition_1}.

\begin{figure}[t]
  \centering
  \subfigure[]{
%  \begin{minipage}[t]{0.5\textwidth}
%  \centering
  \includegraphics[height=0.3\textwidth, width=0.44\textwidth]{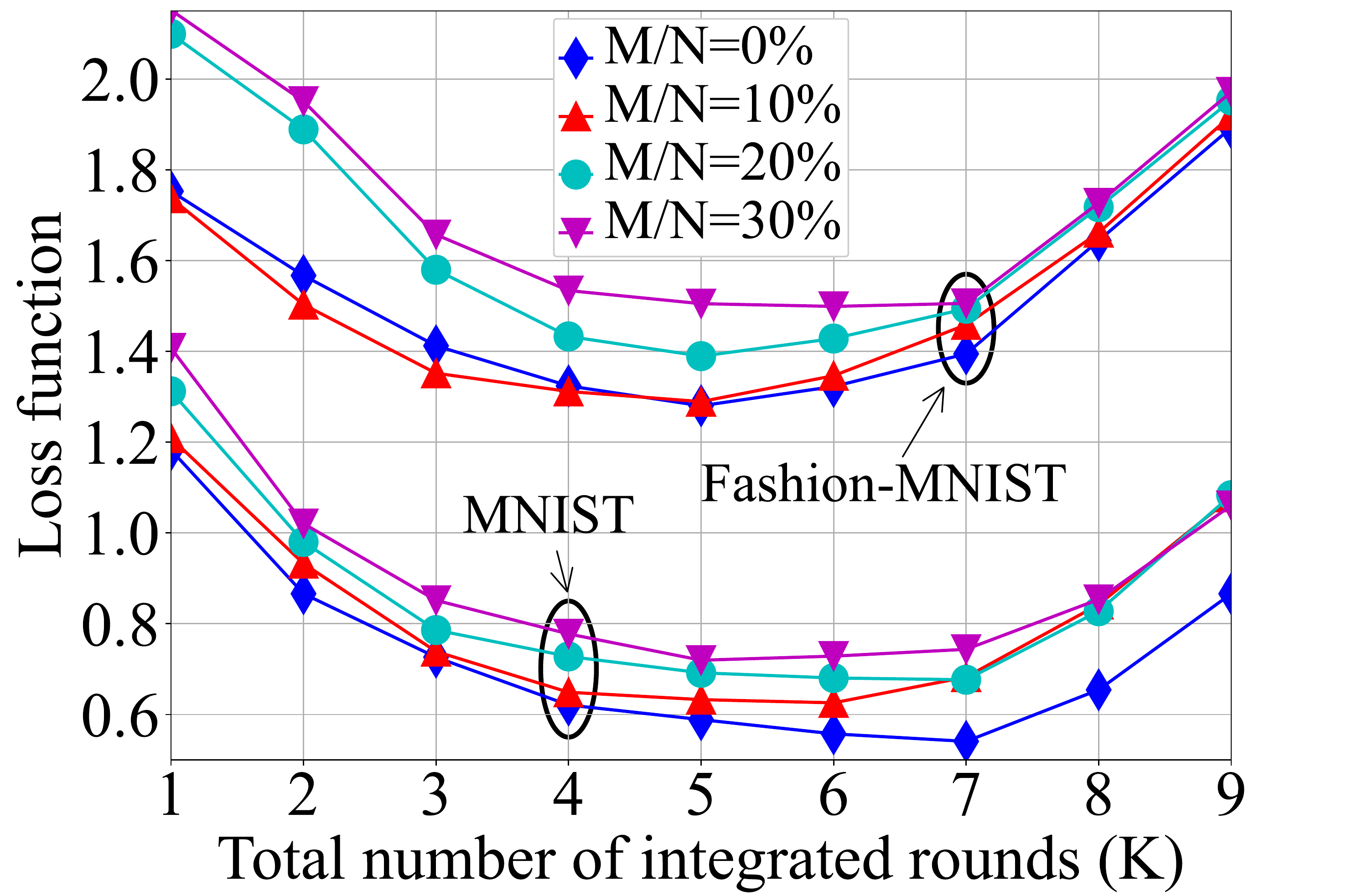}
%  \end{minipage}
  }
  %}
  \subfigure[]{
%  \begin{minipage}[t]{0.5\textwidth}
%  \centering
  \includegraphics[height=0.3\textwidth, width=0.44\textwidth]{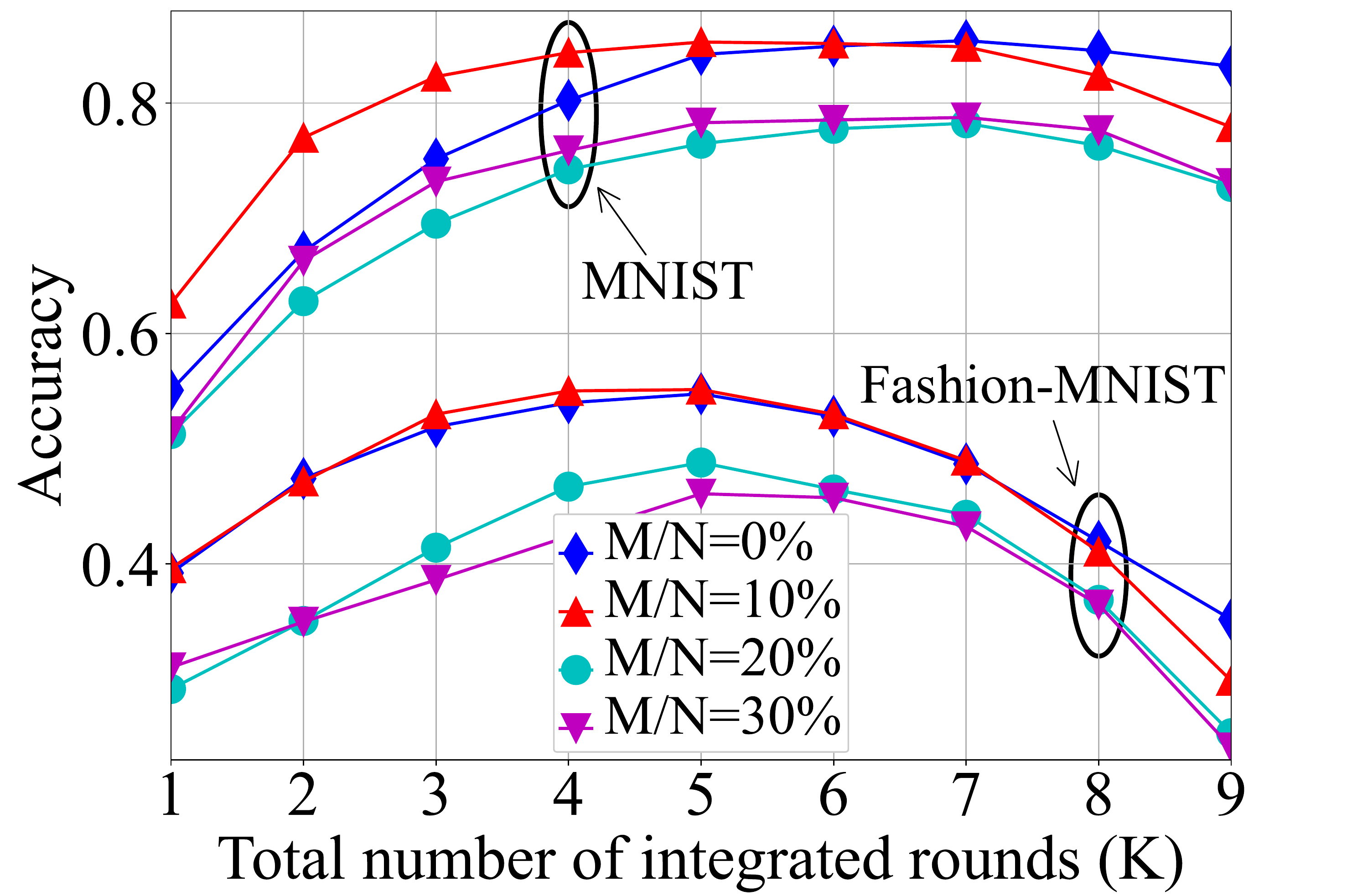}
%  \end{minipage}
  }
  \vspace{-0.3cm}
  \caption{Loss function and accuracy versus $K$ under various $\frac{M}{N}$ for MNIST and Fashion-MNIST.}
  %\vspace{-0.2cm}
    \label{fig_M_loss}
\end{figure}

\begin{table}[t]
\centering
\caption{The optimal training time and corresponding accuracy for different values of $\frac{M}{N}$. }
\vspace{-0.1cm}
\begin{tabular}{c|c|c|c|c}
\hline
\multirow{3}{*}{\tabincell{c}{Lazy \\ ratio}} &  \multicolumn{2}{c|}{$\tau \alpha K^*$}& \multicolumn{2}{c}{Maximal accuracy}\\
\cline{2-5}
&\multirow{2}{*}{MNIST} & Fashion- & \multirow{2}{*}{MNIST}& Fashion-\\
& & MNIST & & MNIST\\
\hline
$\frac{M}{N}=0\%$ & 30 & 50 & 85.53\% & 54.86\% \\
\hline
$\frac{M}{N}=10\%$ & 40 & 50 & 85.33\% & 54.76\% \\
\hline
$\frac{M}{N}=20\%$ & 50 & 80 & 78.11\%  & 48.92\% \\
\hline
$\frac{M}{N}=30\%$ & 50 & 80 & 78.80\% & 46.25\% \\
\hline
\end{tabular}
\vspace{-0.1cm}
\label{tab:M}
\end{table}

Fig.~\ref{fig_5} plots the experimental results of the loss function and accuracy on MNIST and Fashion-MNIST for different values of values of $\eta$, while Table~\ref{tab:eta} illustrates the optimal mining time and corresponding accuracy.
First, from Table~\ref{tab:eta}, we find that the optimal mining time rises as $\eta$ grows, which is in line with Corollary \ref{eta_remark}. For example, using MNIST, $\beta K^*$ rises from $54$ to $72$ as $\eta$ grows from $0.005$ to $0.1$. Second, from Fig.~\ref{fig_5}(a), the loss function drops as $\eta$ increases except $\eta=0.1$. This is because $\eta$ grows significantly when $\eta L<1$, and our developed upper bound is no longer suitable. For example, when $\eta>0.05$, the loss function increases as $\eta$ rises in our experiments for both MNIST and Fashion-MNIST.
\subsection{Experiments on Performance with Lazy Clients}
%In this subsection, we progress the learning task with lazy clients on MNIST and Fashion-MNIST, and show the learning performance and the optimal $K$ on various lazy clients ratio $\frac{M}{N}$ and various power of artificial noise $\sigma^2$.

\begin{figure}[t]
  \centering
  \subfigure[]{
%  \begin{minipage}[t]{0.5\textwidth}
%  \centering
  \includegraphics[height=0.3\textwidth, width=0.44\textwidth]{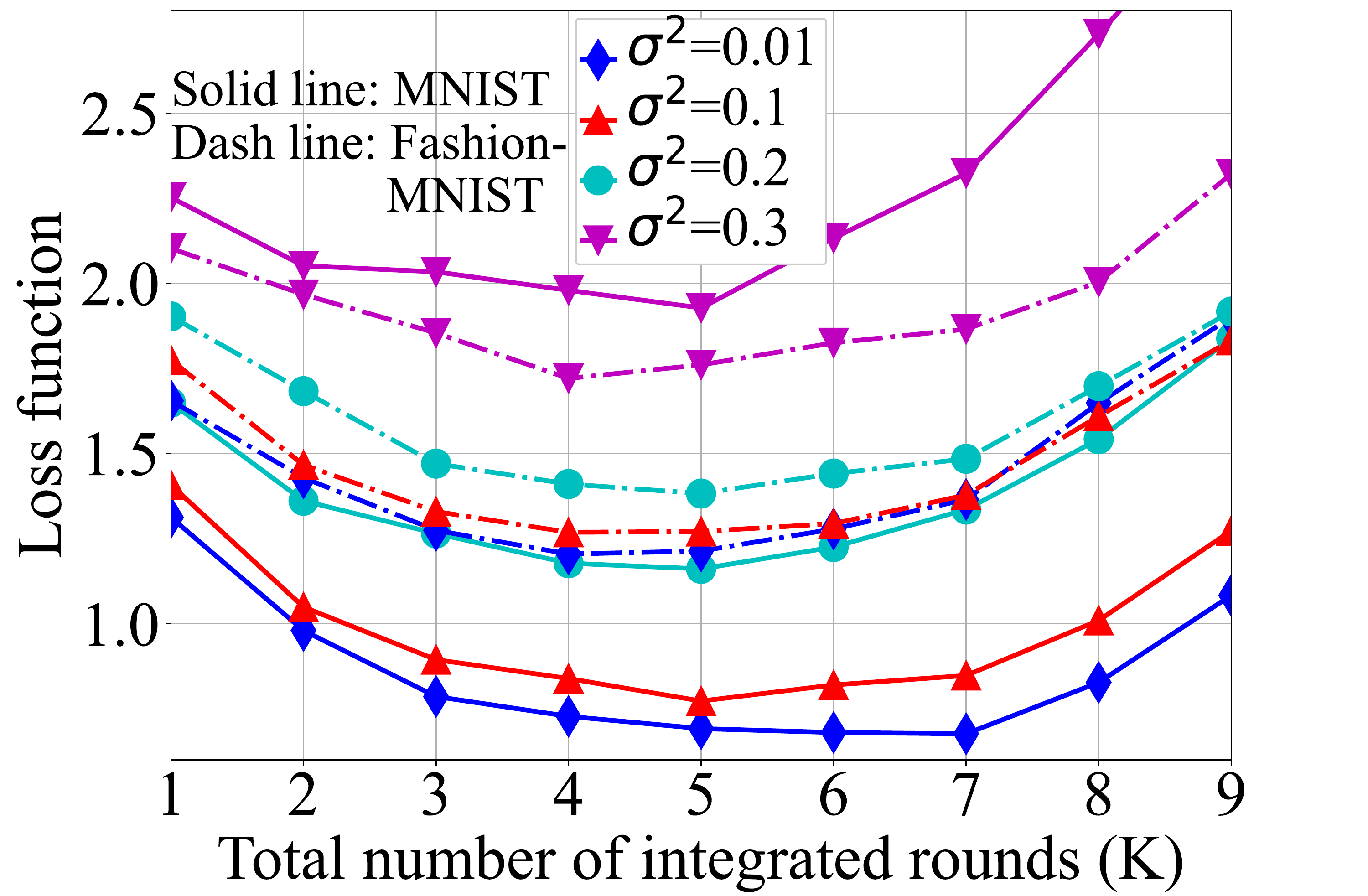}
%  \end{minipage}
  }
  %}
  \subfigure[]{
%  \begin{minipage}[t]{0.5\textwidth}
%  \centering
  \includegraphics[height=0.3\textwidth, width=0.44\textwidth]{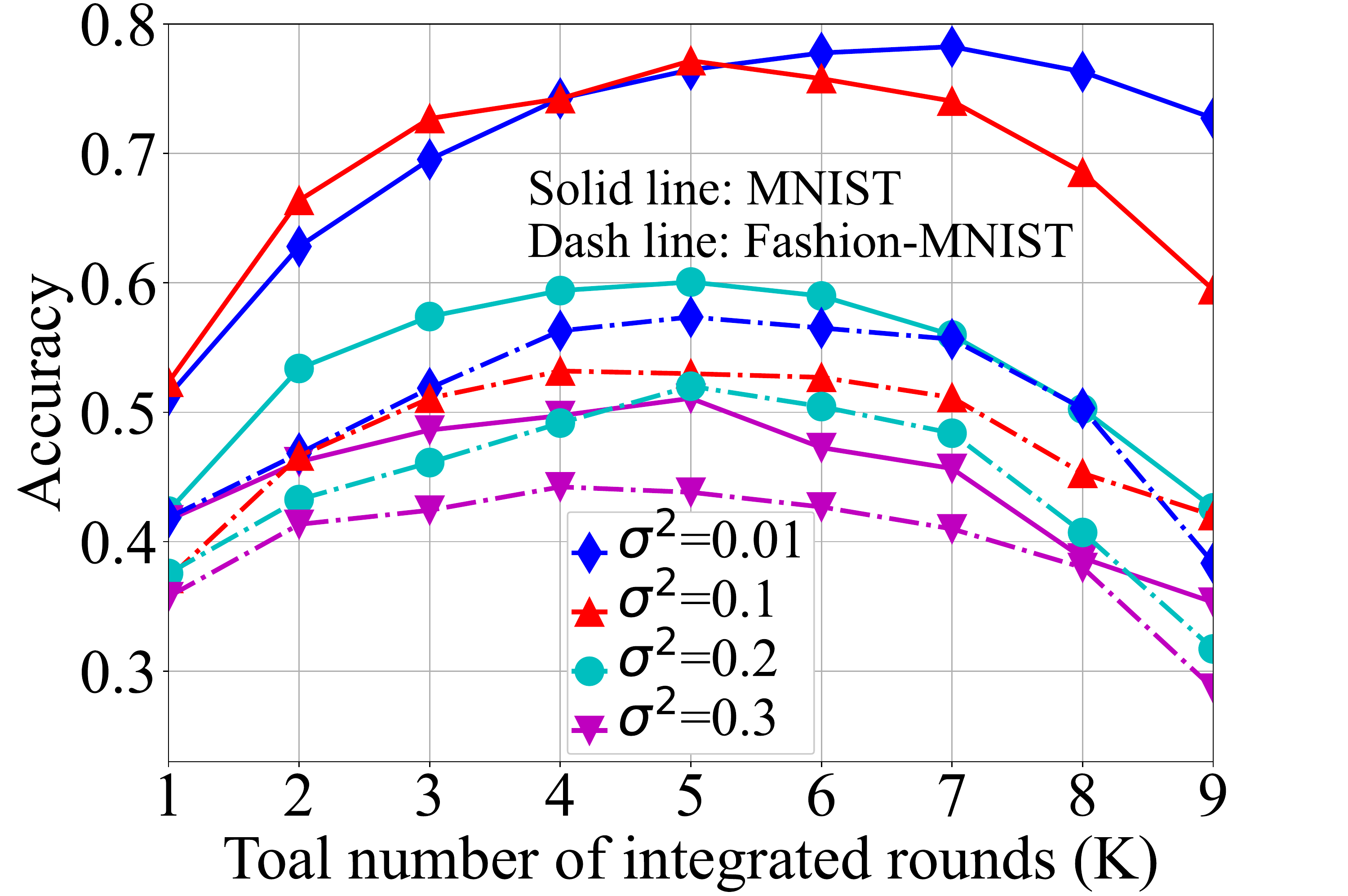}
%  \end{minipage}
  }
  %}
  \vspace{-0.3cm}
  \caption{Loss function and accuracy versus $K$ under various $\sigma^2$ for MNIST and Fashion-MNIST.}
  %\vspace{-0.2cm}
  \label{fig_9}
\end{figure}

\begin{table}[t]
\centering
\caption{The optimal training time and corresponding accuracy for different values of $\sigma^2$. }
\vspace{-0.1cm}
\begin{tabular}{c|c|c|c|c}
\hline
\multirow{3}{*}{\tabincell{c}{Power of \\ artificial \\ noise}} &  \multicolumn{2}{c|}{$\tau \alpha K^*$}& \multicolumn{2}{c}{Maximal accuracy}\\
\cline{2-5}
&\multirow{2}{*}{MNIST} & Fashion- & \multirow{2}{*}{MNIST}& Fashion-\\
& & MNIST & & MNIST\\
\hline
$\sigma^2=0.01$ & 30 & 50 & 78.35\% & 57.44\% \\
\hline
$\sigma^2=0.1$ & 50 & 50 & 77.22\% & 53.19\% \\
\hline
$\sigma^2=0.2$ & 50 & 50 & 59.96\%  & 52.06\% \\
\hline
$\sigma^2=0.3$ & 50 & 60 & 50.94\% & 44.08\% \\
\hline
\end{tabular}
\vspace{0.1cm}
\label{tab:sigma}
\end{table}

Fig.~\ref{fig_M_loss} plots the experimental results of the loss function and accuracy on MNIST and Fashion-MNIST for different values of values of lazy ratio $\frac{M}{N}$, while Table~\ref{tab:M} shows the optimal training time and corresponding accuracy. We set the power of artificial noise $\sigma^2=0.01$.
First, from Table~\ref{tab:M}, it is observed the optimal training time steps up as $\frac{M}{N}$ increases. For example, using MNIST, the time allocated to training rises from $30$ to $50$ as $\frac{M}{N}$ increases from $0\%$ to $30\%$. This observation is consistent with Corollary \ref{pro_lazy_sigma}.
Second, from Fig.~\ref{fig_M_loss}(a), the learning performance degrades as $\frac{M}{N}$ grows. This is because more lazy clients involved in the system as $\frac{M}{N}$ grows, leading to lower training efficiency.
%Third, the optimal $K$ of Fashion-MNIST also follows the trend of learning performance of MNIST.

Fig.~\ref{fig_9} plots the experimental results of the loss function and accuracy on MNIST and Fashion-MNIST for different values of values of $\sigma^2$, while Table~\ref{tab:sigma} shows the optimal training time and corresponding accuracy. We set $\frac{M}{N}=20\%$.
First, from Table~\ref{tab:sigma}, we notice that the optimal training time $\tau \alpha K^*$ grows as $\sigma^2$ increases, which agrees with Corollary \ref{pro_lazy_sigma}. For example, using MNIST, $\tau \alpha K^*$ grows from $30$ to $50$ as $\sigma^2$ increases from $0.01$ to $0.3$.
Second, from Fig.~\ref{fig_9}(a), the learning performance of BLADE-FL (i.e., loss function and accuracy) degrades as the noise power $\sigma^2$ goes larger.

%The experimental results of various $\frac{M}{N}$ and $\sigma^2$ is consistent with \textbf{Corollary \ref{pro_lazy_sigma}}.
\subsection{Experiments of Differential Privacy Mechanism}\label{appendix_expdp}
\begin{figure}[t]%%%%%%%%%%%%%%%%%½ûֹͼƬ¸¡¶¯
  \centering
  % Requires \usepackage{float}À´½ûÖ¹¸¡¶¯
  \includegraphics[height=0.3\textwidth, width=0.4\textwidth]{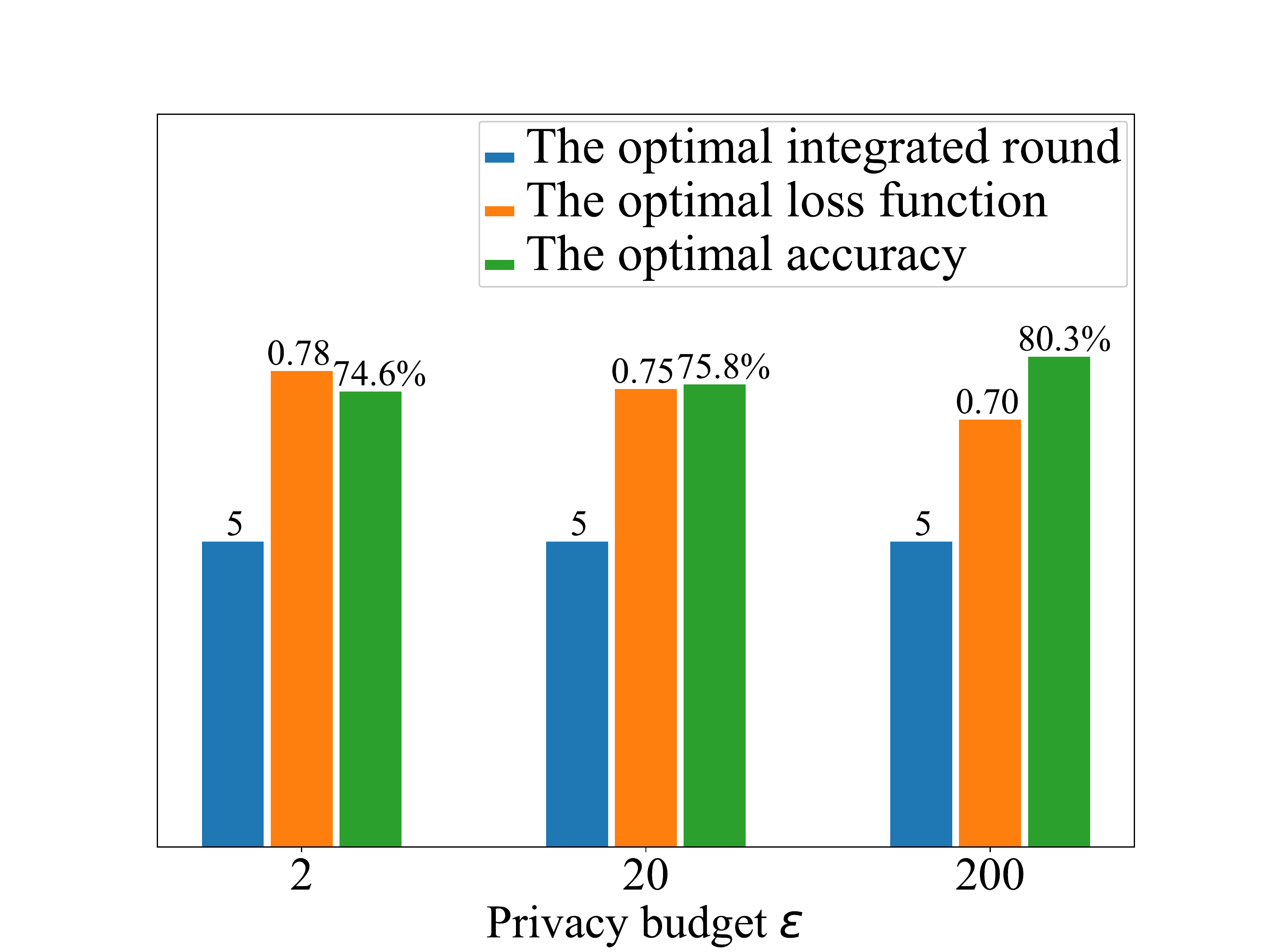}\\
  %\vspace{-0.1cm}
  \caption{The optimal integrated round, loss function, and accuracy versus various privacy budget of MNIST.}
  %\vspace{-0.4cm}
    \label{fig_dp}
\end{figure}

\begin{figure}[t]%%%%%%%%%%%%%%%%%½ûֹͼƬ¸¡¶¯
  \centering
  % Requires \usepackage{float}À´½ûÖ¹¸¡¶¯
  \includegraphics[height=0.3\textwidth, width=0.4\textwidth]{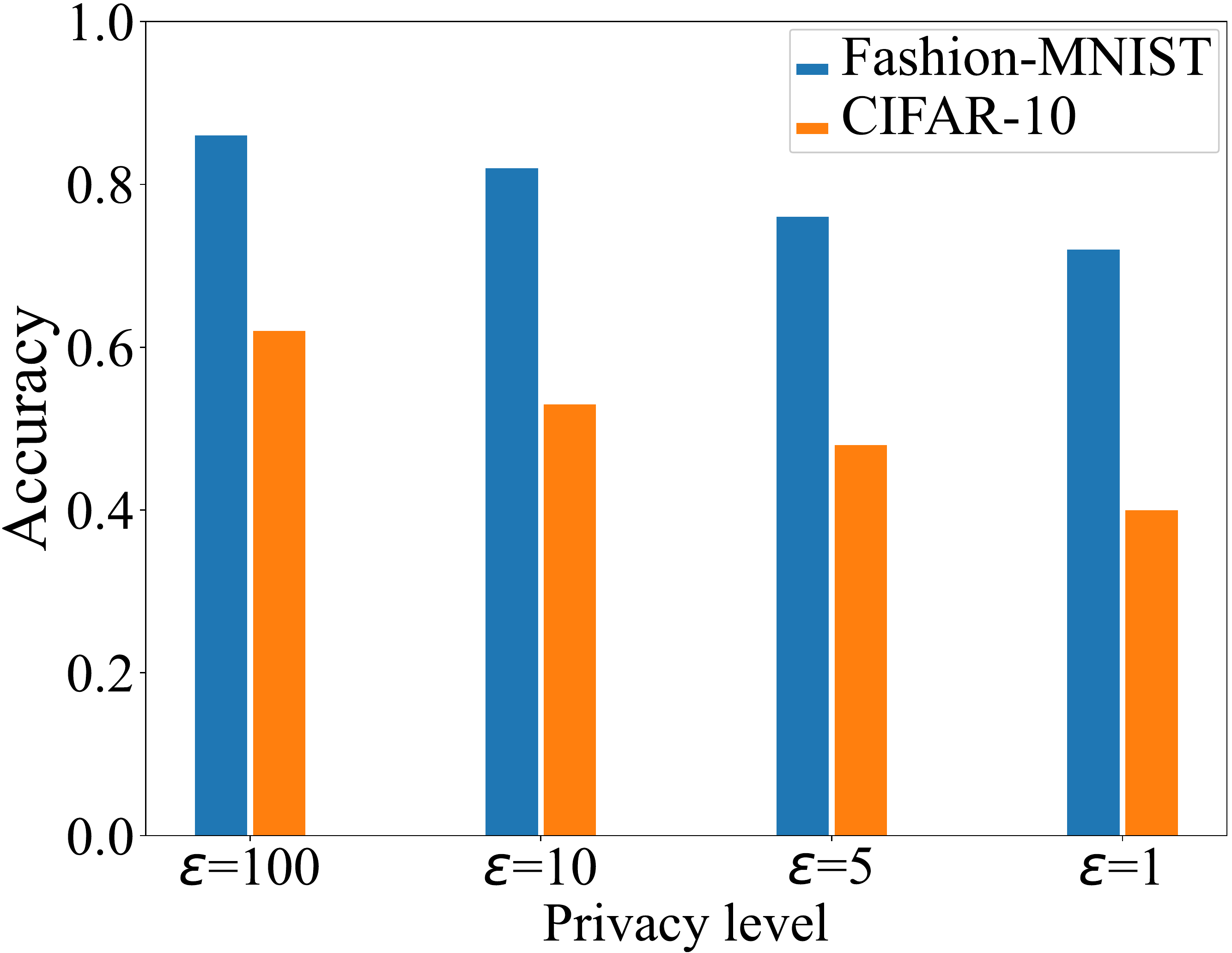}\\
  %\vspace{-0.1cm}
  \caption{The optimal accuracy versus various $\epsilon$ of Fashion-MNIST and Cifar-10.}
  %\vspace{-0.4cm}
    \label{fig_privacy}
\end{figure}

Based on \cite{DBLP:journals/tifs/WeiLDMYFJQP20}, we have the relationship between Gaussian noise variance and the privacy budget $\epsilon$. Then, we apply local differential privacy on each client by adding random Gaussian noises on the uploaded models in each integrated round. We show the experiments of various $\epsilon$ versus the optimal integrated rounds, the optimal loss function and accuracy in Fig.~\ref{fig_dp} and Fig.~\ref{fig_privacy}.

Fig.~\ref{fig_dp} plots the experimental results of the optimal integrated round, loss function and accuracy on MNIST for different values of values of privacy budget $\epsilon$. The $\epsilon$ of differential privacy algorithm represents the privacy budget \cite{DBLP:journals/iotj/ChamikaraBKLCA20} (a smaller $\epsilon$ leads to a higher level of privacy protection). First, we notice that the optimal loss function decreases as $\epsilon$ increases, while the optimal accuracy grows up. That is, the higher privacy budget leads to a lower learning performance. Second, from Fig.~\ref{fig_dp}, we observe that the optimal integrated round is not related to $\epsilon$. In this way, the method of privacy protection does not affect the optimal computing power allocation of BLADE-FL, which is consistent with the discussion of privacy of BLADE-FL in {Section \ref{Sec:privacy}}. Therefore, clients in BLADE-FL can use privacy protection algorithm to meet the privacy needs without changing their own optimal computing power allocation.

The testing accuracies of the Fashion-MNIST and Cifar-10 dataset are plotted in Fig.~\ref{fig_privacy} with respect to different privacy level $\epsilon$. This figure shows that the system achieves a higher performance with a larger value of $\epsilon$. However, a larger $\epsilon$ implies that the system is under a weaker privacy protection.

\section{Conclusions}\label{sec:Concl}
In this paper, we have proposed a BLADE-FL framework that integrates the training and mining process in each client, to overcome the single-point-failure of centralized network and maintain the privacy promoting capabilities of the FL system. In order to evaluate the learning performance of BLADE-FL, we have developed an upper bound on the loss function. Also, we have verified that the upper bound is convex with respect to the total number of integrated rounds $K$ and have minimized the upper bound by optimizing $K$.
Moreover, we have investigated a unique problem in the proposed BLADE-FL system, called the lazy client problem and have derived an upper bound on the loss function with lazy clients. We have included experimental results, which have been seen to be consistent with the analytical results. In particular, the developed upper bound is close to the experimental results (e.g., the gap can be lower than $5\%$), and the optimal $K$ that minimizes the upper bound also reaches the minimum of the loss function in the experimental results.

A few directions of future works are discussed as follows.
First, it is of interest to design an efficient incentive mechanism for BLADE-FL that encourages all clients to actively follow the protocol.
Second, another challenging issue of BLADE-FL is to detect plagiarism behaviors and discourage lazy clients. The detection of lazy clients will be addressed in our future work.
%Third, due to the Heterogeneity of computing and communication resources and the energy constraints, the performance analysis and the optimization of learning performance on asynchronous or heterogenous BLADE-FL is for further study.

%/newpage

% or

\ifCLASSOPTIONcompsoc
%\else
%  % regular IEEE prefers the singular form
%\fi


%\newpage
%\appendices
%\section{Proof of Theorem1}
%
%
%\section{Proof of Theorem~2}
%
%
%
%\section{Proof of Remark \ref{delta_remark}}
%
%
%\section{Proof of Corollary \ref{eta_remark}}
%
%\section{Proof of Theorem~3}
%
%
%
%\section{Proof of Corollary \ref{pro_lazy_sigma}}

\ifCLASSOPTIONcaptionsoff
  \newpage
\fi

\bibliographystyle{IEEEtran}
\bibliography{reference}

\begin{thebibliography}{10}
\providecommand{\url}[1]{#1}
\csname url@samestyle\endcsname
\providecommand{\newblock}{\relax}
\providecommand{\bibinfo}[2]{#2}
\providecommand{\BIBentrySTDinterwordspacing}{\spaceskip=0pt\relax}
\providecommand{\BIBentryALTinterwordstretchfactor}{4}
\providecommand{\BIBentryALTinterwordspacing}{\spaceskip=\fontdimen2\font plus
\BIBentryALTinterwordstretchfactor\fontdimen3\font minus
  \fontdimen4\font\relax}
\providecommand{\BIBforeignlanguage}[2]{{%
\expandafter\ifx\csname l@#1\endcsname\relax
\typeout{** WARNING: IEEEtran.bst: No hyphenation pattern has been}%
\typeout{** loaded for the language `#1'. Using the pattern for}%
\typeout{** the default language instead.}%
\else
\language=\csname l@#1\endcsname
\fi
#2}}
\providecommand{\BIBdecl}{\relax}
\BIBdecl

\bibitem{TheNextGrandChallenges}
R.~{Ranjan}, O.~{Rana}, S.~{Nepal}, M.~{Yousif}, P.~{James}, Z.~{Wen},
  S.~{Barr}, P.~{Watson}, P.~P. {Jayaraman}, D.~{Georgakopoulos}, M.~{Villari},
  M.~{Fazio}, S.~{Garg}, R.~{Buyya}, L.~{Wang}, A.~Y. {Zomaya}, and
  S.~{Dustdar}, ``The next grand challenges: Integrating the internet of things
  and data science,'' \emph{IEEE Trans. Cloud Comput.}, vol.~5, no.~3, pp.
  12--26, Jun. 2018.

\bibitem{8805879}
X.~{Li}, K.~{Li}, D.~{Qiao}, Y.~{Ding}, and D.~{Wei}, ``Application research of
  machine learning method based on distributed cluster in information
  retrieval,'' in \emph{Proceedings of the International Conference on
  Communications, Information System and Computer Engineering (CISCE)}, Haikou,
  Hainan, China, Jul. 2019.

\bibitem{DBLP:conf/aistats/McMahanMRHA17}
B.~McMahan, E.~Moore, D.~Ramage, S.~Hampson, and B.~A. Arcas,
  ``Communication-efficient learning of deep networks from decentralized
  data,'' in \emph{Proceedings of the 20th International Conference on
  Artificial Intelligence and Statistics, {AISTATS}}, ser. Proceedings of
  Machine Learning Research (PMLR), vol.~54, Fort Lauderdale, FL, May 2017, pp.
  1273--1282.

\bibitem{9090973}
S.~{Shaham}, M.~{Ding}, B.~{Liu}, S.~{Dang}, Z.~{Lin}, and J.~{Li}, ``Privacy
  preservation in location-based services: A novel metric and attack model,''
  \emph{IEEE Trans. Mobile Comput.}, Early access, 2020.

\bibitem{DBLP:journals/corr/KonecnyMRR16}
\BIBentryALTinterwordspacing
J.~Konecn{\'{y}}, H.~B. McMahan, D.~Ramage, and P.~Richt{\'{a}}rik, ``Federated
  optimization: Distributed machine learning for on-device intelligence.''
  [Online]. Available: \url{http://arxiv.org/abs/1610.02527}
\BIBentrySTDinterwordspacing

\bibitem{8951246}
S.~{Shaham}, M.~{Ding}, B.~{Liu}, S.~{Dang}, Z.~{Lin}, and J.~{Li}, ``Privacy
  preserving location data publishing: A machine learning approach,''
  \emph{IEEE Trans. Knowl. Data Eng.}, Early access, 2020.

\bibitem{9048613}
C.~{Ma}, J.~{Li}, M.~{Ding}, H.~H. {Yang}, F.~{Shu}, T.~Q.~S. {Quek}, and H.~V.
  {Poor}, ``On safeguarding privacy and security in the framework of federated
  learning,'' \emph{IEEE Netw.}, vol.~34, no.~4, pp. 242--248, Aug. 2020.

\bibitem{DBLP:journals/corr/abs-2007-02056}
\BIBentryALTinterwordspacing
C.~Ma, J.~Li, M.~Ding, B.~Liu, K.~Wei, J.~Weng, and H.~V. Poor, ``{RDP-GAN:}
  {A} r{\'{e}}nyi-differential privacy based generative adversarial network,''
  2020. [Online]. Available: \url{https://arxiv.org/abs/2007.02056}
\BIBentrySTDinterwordspacing

\bibitem{DBLP:journals/tifs/WeiLDMYFJQP20}
K.~Wei, J.~Li, M.~Ding, C.~Ma, H.~H. Yang, F.~Farokhi, S.~Jin, T.~Q.~S. Quek,
  and H.~V. Poor, ``Federated learning with differential privacy: Algorithms
  and performance analysis,'' \emph{{IEEE} Trans. Inf. Forensics Secur.},
  vol.~15, no.~1, pp. 3454--3469, Apr. 2020.

\bibitem{DBLP:journals/tist/YangLCT19}
Q.~Yang, Y.~Liu, T.~Chen, and Y.~Tong, ``Federated machine learning: Concept
  and applications,'' \emph{{ACM} Trans. Intell. Syst. Technol.}, vol.~10,
  no.~2, pp. 1--19, Feb. 2019.

\bibitem{DBLP:journals/corr/abs-1912-01218}
\BIBentryALTinterwordspacing
D.~van Esch, E.~Sarbar, T.~Lucassen, J.~O'Brien, T.~Breiner, M.~Prasad,
  E.~Crew, C.~Nguyen, and F.~Beaufays, ``Writing across the world's languages:
  Deep internationalization for gboard, the google keyboard.'' [Online].
  Available: \url{http://arxiv.org/abs/1912.01218}
\BIBentrySTDinterwordspacing

\bibitem{nakamoto2008peer}
\BIBentryALTinterwordspacing
S.~Nakamoto, ``Bitcoin: A peer-to-peer electronic cash system.'' [Online].
  Available: \url{https://bitcoin.org/bitcoin.pdf}
\BIBentrySTDinterwordspacing

\bibitem{DBLP:journals/fgcs/ReynaMCSD18}
A.~Reyna, C.~Mart{\'{\i}}n, J.~Chen, E.~Soler, and M.~D{\'{\i}}az, ``On
  blockchain and its integration with $\mathrm{IoT}$. challenges and
  opportunities,'' \emph{Future Gener. Comput. Syst.}, vol.~88, no.~1, pp.
  173--190, Nov. 2018.

\bibitem{8436042}
Z.~{Xiong}, Y.~{Zhang}, D.~{Niyato}, P.~{Wang}, and Z.~{Han}, ``When mobile
  blockchain meets edge computing,'' \emph{IEEE Commun. Mag.}, vol.~56, no.~8,
  pp. 33--39, Aug. 2018.

\bibitem{DBLP:journals/corr/abs-1808-03949}
\BIBentryALTinterwordspacing
H.~Kim, J.~Park, M.~Bennis, and S.~Kim, ``On-device federated learning via
  blockchain and its latency analysis.'' [Online]. Available:
  \url{http://arxiv.org/abs/1808.03949}
\BIBentrySTDinterwordspacing

\bibitem{9347812}
X.~Qu, S.~Wang, Q.~Hu, and X.~Cheng, ``Proof of federated learning: A novel
  energy-recycling consensus algorithm,'' \emph{{IEEE} Trans. Parallel
  Distributed Syst.}, vol.~32, no.~8, pp. 2074--2085, Feb. 2021.

\bibitem{9399813}
L.~Feng, Y.~Zhao, S.~Guo, X.~Qiu, W.~Li, and P.~Yu, ``Blockchain-based
  asynchronous federated learning for internet of things,'' \emph{IEEE
  Transactions on Computers}, Early Access, 2021.

\bibitem{9272656}
M.~P. Uddin, Y.~Xiang, X.~Lu, J.~Yearwood, and L.~Gao, ``Mutual information
  driven federated learning,'' \emph{{IEEE} Trans. Parallel Distributed Syst.},
  vol.~32, no.~7, pp. 1526--1538, Nov. 2020.

\bibitem{8470083}
Z.~{Xiong}, S.~{Feng}, W.~{Wang}, D.~{Niyato}, P.~{Wang}, and Z.~{Han},
  ``Cloud/fog computing resource management and pricing for blockchain
  networks,'' \emph{IEEE Internet of Things Journal}, vol.~6, no.~3, pp.
  4585--4600, Sep. 2019.

\bibitem{DBLP:journals/tii/LuHDMZ20a}
Y.~Lu, X.~Huang, Y.~Dai, S.~Maharjan, and Y.~Zhang, ``Blockchain and federated
  learning for privacy-preserved data sharing in industrial $\mathrm{IoT}$,''
  \emph{{IEEE} Trans. Ind. Informatics}, vol.~16, no.~6, pp. 4177--4186, Jun.
  2020.

\bibitem{DBLP:journals/wc/LiuPKINE20}
Y.~Liu, J.~Peng, J.~Kang, A.~M. Iliyasu, D.~Niyato, and A.~A.~A. El{-}Latif,
  ``A secure federated learning framework for 5g networks,'' \emph{{IEEE}
  Wirel. Commun.}, vol.~27, no.~4, pp. 24--31, Aug. 2020.

\bibitem{9347025}
Y.~Chen, Q.~Chen, and Y.~Xie, ``A methodology for high-efficient
  federated-learning with consortium blockchain,'' in \emph{IEEE 4th Conference
  on Energy Internet and Energy System Integration (EI2), Wuhan, China}, Oct.
  2020, pp. 3090--3095.

\bibitem{DBLP:journals/tpds/ShayanFYB21}
M.~Shayan, C.~Fung, C.~J.~M. Yoon, and I.~Beschastnikh, ``Biscotti: {A}
  blockchain system for private and secure federated learning,'' \emph{{IEEE}
  Trans. Parallel Distributed Syst.}, vol.~32, no.~7, pp. 1513--1525, Dec.
  2020.

\bibitem{DBLP:conf/wcnc/Pokhrel020}
S.~R. Pokhrel and J.~Choi, ``A decentralized federated learning approach for
  connected autonomous vehicles,'' in \emph{{IEEE} Wireless Communications and
  Networking Conference Workshops, {WCNC} Workshops 2020, Seoul, Korea (South),
  April 6-9, 2020}.

\bibitem{DBLP:journals/access/ToyodaZZM20}
K.~Toyoda, J.~Zhao, A.~N. Zhang, and P.~T. Mathiopoulos, ``Blockchain-enabled
  federated learning with mechanism design,'' \emph{{IEEE} Access}, vol.~8, pp.
  219\,744--219\,756, 2020.

\bibitem{DBLP:journals/network/LuHZMZ21}
Y.~Lu, X.~Huang, K.~Zhang, S.~Maharjan, and Y.~Zhang, ``Blockchain and
  federated learning for 5g beyond,'' \emph{{IEEE} Netw.}, vol.~35, no.~1, pp.
  219--225, Dec. 2021.

\bibitem{9357330}
S.~R. Pokhrel, ``Blockchain brings trust to collaborative drones and leo
  satellites: An intelligent decentralized learning in the space,'' \emph{IEEE
  Sensors Journal}, Early Access, 2021.

\bibitem{DBLP:journals/iotj/ZhangLYLLLCXZ21}
W.~Zhang, Q.~Lu, Q.~Yu, Z.~Li, Y.~Liu, S.~K. Lo, S.~Chen, X.~Xu, and L.~Zhu,
  ``Blockchain-based federated learning for device failure detection in
  industrial iot,'' \emph{{IEEE} Internet Things J.}, vol.~8, no.~7, pp.
  5926--5937, Oct. 2020.

\bibitem{8946151}
J.~{Li}, T.~{Liu}, D.~{Niyato}, P.~{Wang}, J.~{Li}, and Z.~{Han},
  ``Contract-based approach for security deposit in blockchain networks with
  shards,'' in \emph{Proceedings of the 2019 IEEE International Conference on
  Blockchain (Blockchain)}, Atlanta, USA, Jul. 2019, pp. 75--82.

\bibitem{DBLP:journals/corr/abs-1907-09693}
\BIBentryALTinterwordspacing
Q.~Li, Z.~Wen, and B.~He, ``Federated learning systems: Vision, hype and
  reality for data privacy and protection.'' [Online]. Available:
  \url{http://arxiv.org/abs/1907.09693}
\BIBentrySTDinterwordspacing

\bibitem{DBLP:journals/sigops/DemersGHILSSST88}
A.~J. Demers, D.~H. Greene, C.~Hauser, W.~Irish, J.~Larson, S.~Shenker, H.~E.
  Sturgis, D.~C. Swinehart, and D.~B. Terry, ``Epidemic algorithms for
  replicated database maintenance,'' \emph{{ACM} {SIGOPS} Oper. Syst. Rev.},
  vol.~22, no.~1, pp. 8--32, Jan. 1988.

\bibitem{DBLP:journals/cem/PuthalMMKD18}
D.~Puthal, N.~Malik, S.~P. Mohanty, E.~Kougianos, and G.~Das, ``Everything you
  wanted to know about the blockchain: Its promise, components, processes, and
  problems,'' \emph{{IEEE} Consumer Electron. Mag.}, vol.~7, no.~4, pp. 6--14,
  Jul. 2018.

\bibitem{DBLP:journals/cacm/EyalS18}
I.~Eyal and E.~G. Sirer, ``Majority is not enough: bitcoin mining is
  vulnerable,'' \emph{{ACM} Commun.}, vol.~61, no.~7, pp. 95--102, Jul. 2018.

\bibitem{DBLP:journals/tpds/XuWLGLYG19}
C.~Xu, K.~Wang, P.~Li, S.~Guo, J.~Luo, B.~Ye, and M.~Guo, ``Making big data
  open in edges: {A} resource-efficient blockchain-based approach,''
  \emph{{IEEE} Trans. Parallel Distributed Syst.}, vol.~30, no.~4, pp.
  870--882, Apr. 2019.

\bibitem{9242286}
X.~{Deng}, J.~{Li}, L.~{Shi}, Z.~{Wei}, X.~{Zhou}, and J.~{Yuan}, ``Wireless
  powered mobile edge computing: Dynamic resource allocation and throughput
  maximization,'' \emph{IEEE Trans. Mobile Comput.}, Early access, 2020.

\bibitem{DBLP:journals/jsac/WangTSLMHC19}
S.~Wang, T.~Tuor, T.~Salonidis, K.~K. Leung, C.~Makaya, T.~He, and K.~Chan,
  ``Adaptive federated learning in resource constrained edge computing
  systems,'' \emph{{IEEE} J. Sel. Areas Commun.}, vol.~37, no.~6, pp.
  1205--1221, Jun. 2019.

\bibitem{DBLP:journals/corr/abs-2010-05958}
\BIBentryALTinterwordspacing
Z.~Chai, Y.~Chen, L.~Zhao, Y.~Cheng, and H.~Rangwala, ``Fedat: {A}
  communication-efficient federated learning method with asynchronous tiers
  under non-iid data,'' 2020. [Online]. Available:
  \url{https://arxiv.org/abs/2010.05958}
\BIBentrySTDinterwordspacing

\bibitem{DBLP:conf/pkdd/KarimiNS16}
H.~Karimi, J.~Nutini, and M.~Schmidt, ``Linear convergence of gradient and
  proximal-gradient methods under the polyak-{\l}ojasiewicz condition,'' in
  \emph{Proceedings of the Machine Learning and Knowledge Discovery in
  Databases - European Conference, {ECML} {PKDD}}, ser. Lecture Notes in
  Computer Science, vol. 9851.\hskip 1em plus 0.5em minus 0.4em\relax Riva del
  Garda, Italy: Springer, 2016, pp. 795--811.

\bibitem{9119406}
M.~{Tahir}, M.~H. {Habaebi}, M.~{Dabbagh}, A.~{Mughees}, A.~{Ahad}, and K.~I.
  {Ahmed}, ``A review on application of blockchain in 5g and beyond networks:
  Taxonomy, field-trials, challenges and opportunities,'' \emph{IEEE Access},
  vol.~8, pp. 115\,876--115\,904, 2020.

\bibitem{DBLP:conf/trustbus/AbramsonHPPB20}
W.~Abramson, A.~J. Hall, P.~Papadopoulos, N.~Pitropakis, and W.~J. Buchanan,
  ``A distributed trust framework for privacy-preserving machine learning,'' in
  \emph{Proceedings of the Trust, Privacy and Security in Digital Business -
  17th International TrustBus Conference}, ser. Lecture Notes in Computer
  Science, vol. 12395.\hskip 1em plus 0.5em minus 0.4em\relax Bratislava,
  Slovakia: Springer, 2020, pp. 205--220.

\bibitem{DBLP:journals/iotj/ChamikaraBKLCA20}
M.~A.~P. Chamikara, P.~Bert{\'{o}}k, I.~Khalil, D.~Liu, S.~Camtepe, and
  M.~Atiquzzaman, ``Local differential privacy for deep learning,''
  \emph{{IEEE} Internet Things J.}, vol.~7, no.~7, pp. 5827--5842, Nov. 2019.

\bibitem{9055478}
Z.~Li, V.~Sharma, and S.~P.~Mohanty, ``Preserving data privacy via federated
  learning: Challenges and solutions,'' \emph{IEEE Consumer Electronics
  Magazine}, vol.~9, no.~3, pp. 8--16, Apr. 2020.

\bibitem{8919319}
C.~Wu, F.~Zhang, and F.~Wu, ``Distributed modelling approaches for data privacy
  preserving,'' in \emph{IEEE Fifth International Conference on Multimedia Big
  Data (BigMM), Los Angeles, USA}, 2019.

\end{thebibliography}

\begin{IEEEbiography}[{\includegraphics[width=1in,height=1.25in,clip,keepaspectratio]{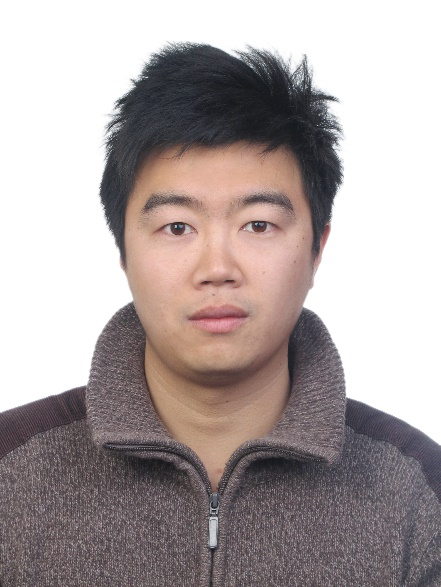}}]
{Jun Li} (M'09-SM'16) received Ph. D degree in Electronic Engineering from Shanghai Jiao Tong University, Shanghai, P. R. China in 2009. From January 2009 to June 2009, he worked in the Department of Research and Innovation, Alcatel Lucent Shanghai Bell as a Research Scientist. From June 2009 to April 2012, he was a Postdoctoral Fellow at the School of Electrical Engineering and Telecommunications, the University of New South Wales, Australia. From April 2012 to June 2015, he is a Research Fellow at the School of Electrical Engineering, the University of Sydney, Australia. From June 2015 to now, he is a Professor at the School of Electronic and Optical Engineering, Nanjing University of Science and Technology, Nanjing, China. He was a visiting professor at Princeton University from 2018 to 2019. His research interests include network information theory, game theory, distributed intelligence, multiple agent reinforcement learning, and their applications in ultra-dense wireless networks, mobile edge computing, network privacy and security, and industrial Internet of things. He has co-authored more than 200 papers in IEEE journals and conferences, and holds 1 US patents and more than 10 Chinese patents in these areas. He was serving as an editor of IEEE Communication Letters and TPC member for several flagship IEEE conferences. He received Exemplary Reviewer of IEEE Transactions on Communications in 2018, and best paper award from IEEE International Conference on 5G for Future Wireless Networks in 2017.
\end{IEEEbiography}

\begin{IEEEbiography}[{\includegraphics[width=1in,height=1.25in,clip,keepaspectratio]{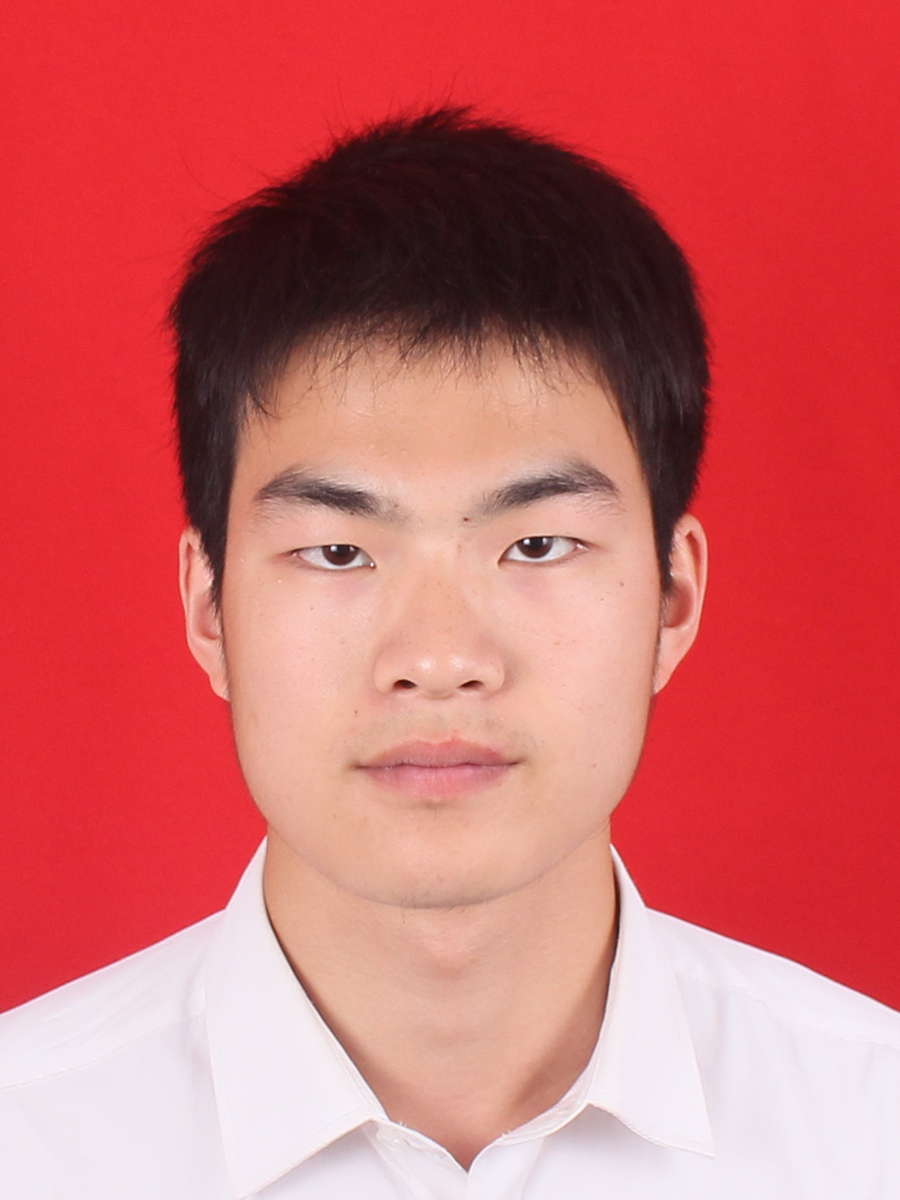}}]
{Yumeng Shao} received the B.S. degree from the School of Electronic and
Optical Engineering, Nanjing University of Science and Technology, Nanjing, China, in 2019, where he is pursuing the M.S. degree currently. His research interests include distributed machine learning, blockchain, game theory, and trusted AI.
\end{IEEEbiography}

\begin{IEEEbiography}[{\includegraphics[width=1in,height=1.25in,clip,keepaspectratio]{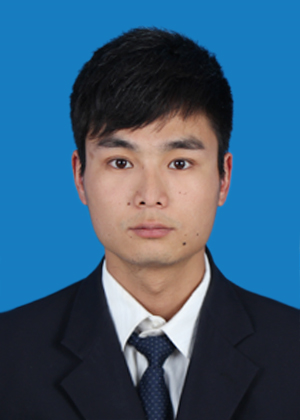}}]
{Kang Wei} received the B.Sc. degree in information engineering from Xidian University, Xi¡¯an, China, in 2014, and the M.Sc. degree from the School of Electronic and
Optical Engineering, Nanjing University of Science and Technology, Nanjing, China, in 2018, where he is currently pursuing the Ph.D. degree. His current research interests include data privacy and security, differential privacy, AI and machine learning, information theory, and channel coding theory in NAND flash memory.
\end{IEEEbiography}

\begin{IEEEbiography}[{\includegraphics[width=1in,height=1.25in,clip,keepaspectratio]{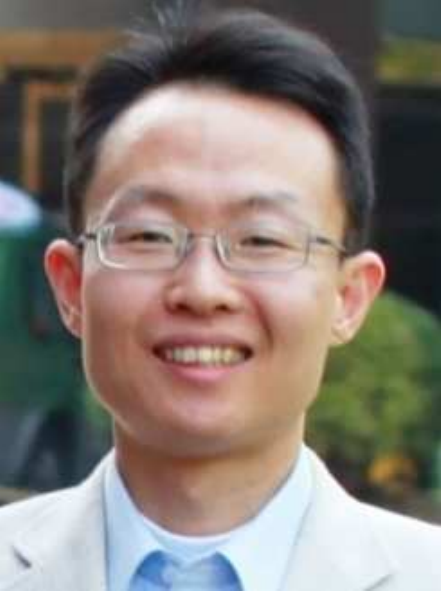}}]
{Ming Ding} (M'12-SM'17) received the B.S. and M.S. degrees (with first class Hons.) in electronics engineering from Shanghai Jiao Tong University (SJTU), Shanghai, China, and the Doctor of Philosophy (Ph.D.) degree in signal and information processing from SJTU, in 2004, 2007, and 2011, respectively. From April 2007 to September 2014, he worked at Sharp Laboratories of China in Shanghai, China as a Researcher/Senior Researcher/Principal Researcher. He also served as the Algorithm Design Director and Programming Director for a system-level simulator of future telecommunication networks in Sharp Laboratories of China for more than 7 years. Currently, he is a senior research scientist at Data61, CSIRO, in Sydney, NSW, Australia. His research interests include information technology, data privacy and security, machine Learning and AI, etc. He has authored over 100 papers in IEEE journals and conferences, all in recognized venues, and around 20 3GPP standardization contributions, as well as a Springer book ``Multi-point Cooperative Communication Systems: Theory and Applications''. Also, he holds 21 US patents and co-invented another 100+ patents on 4G/5G technologies in CN, JP, KR, EU, etc. Currently, he is an editor of IEEE Transactions on Wireless Communications and IEEE Wireless Communications Letters. Besides, he is or has been Guest Editor/Co-Chair/Co-Tutor/TPC member of several IEEE top-tier journals/conferences, e.g., the IEEE Journal on Selected Areas in Communications, the IEEE Communications Magazine, and the IEEE Globecom Workshops, etc. He was the lead speaker of the industrial presentation on unmanned aerial vehicles in IEEE Globecom 2017, which was awarded as the Most Attended Industry Program in the conference. Also, he was awarded in 2017 as the Exemplary Reviewer for IEEE Transactions on Wireless Communications.
\end{IEEEbiography}

\begin{IEEEbiography}[{\includegraphics[width=1in,height=1.25in,clip,keepaspectratio]{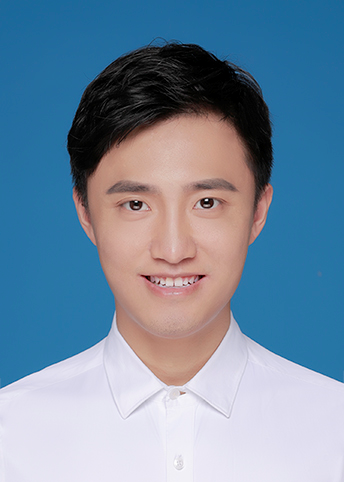}}]
{Chuan Ma} received the B.S. degree from the Beijing University of Posts and Telecommunications, Beijing, China, in 2013 and Ph.D. degree  from the University of Sydney, Australia, in 2018. He is now working as a lecturer at the School of Electronic and Optical Engineering, Nanjing University of Science and Technology, Nanjing, China. He has published more than 10 journal and conference papers, including a best paper in WCNC 2018. His research interests include stochastic geometry, wireless caching networks and machine learning, and now focuses on the big data analysis and privacy preservation.
\end{IEEEbiography}

\begin{IEEEbiography}[{\includegraphics[width=1in,height=1.25in,clip,keepaspectratio]{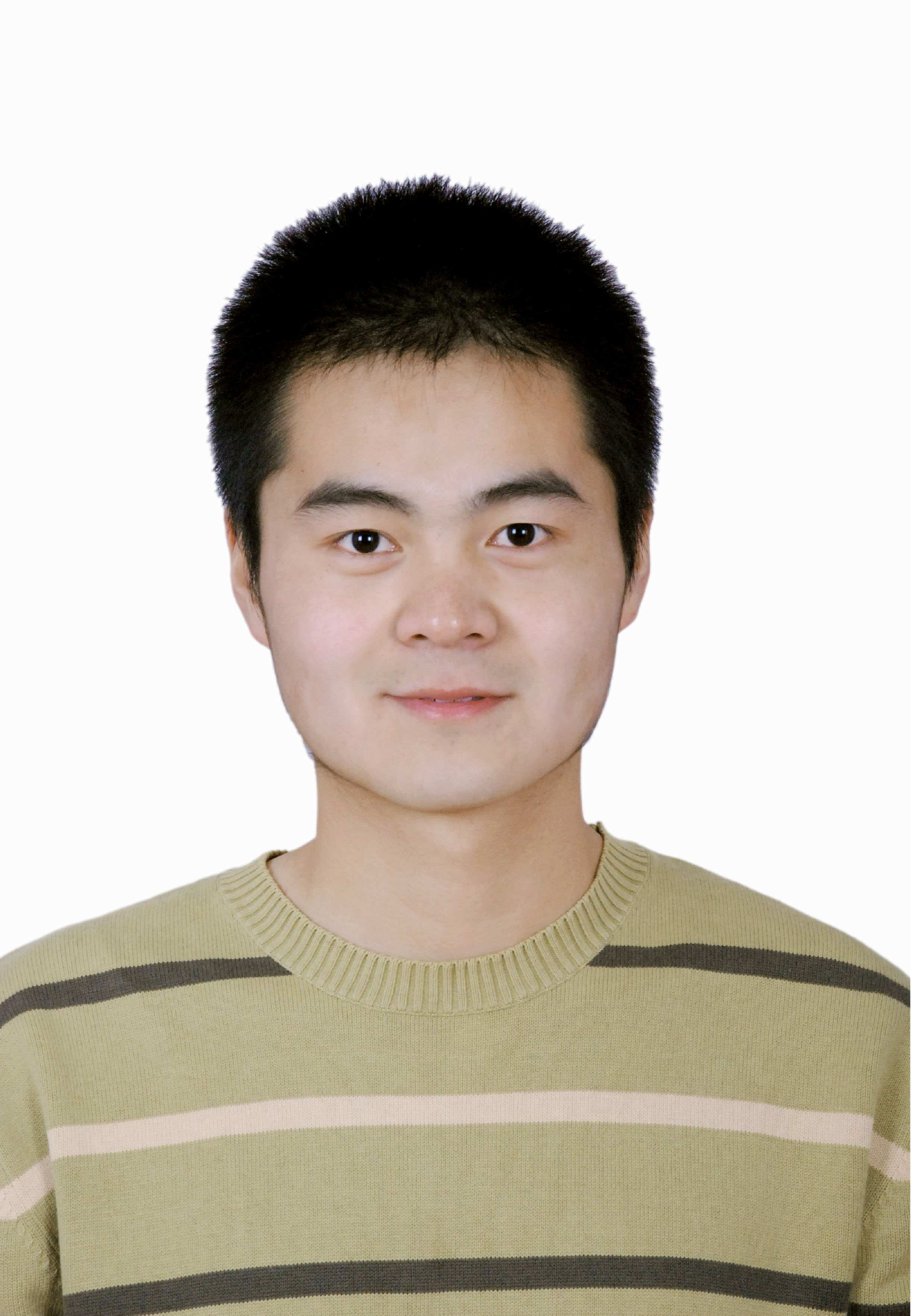}}]
{Long Shi} (S'10-M'15) received the Ph.D. degree in Electrical Engineering from the University of New South Wales, Sydney, Australia, in 2012. From 2013 to 2016, he was a Postdoctoral Fellow at the Institute of Network Coding, Chinese University of Hong Kong, China. From 2014 to 2017, he was a Lecturer at Nanjing University of Aeronautics and Astronautics, Nanjing, China. From 2017 to 2020, he was a Research Fellow at the Singapore University of Technology and Design. Now he is a Professor at the School of Electronic and Optical Engineering, Nanjing University of Science and Technology, Nanjing, China.
\end{IEEEbiography}

\begin{IEEEbiography}[{\includegraphics[width=1in,height=1.25in,clip,keepaspectratio]{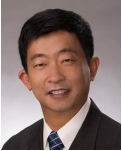}}]
{Zhu Han} received the B.S. degree in electronic engineering from Tsinghua University, in 1997, and the M.S. and Ph.D. degrees in electrical and computer engineering from the University of Maryland, College Park, in 1999 and 2003, respectively. From 2000 to 2002, he was a Research and Development Engineer with JDSU, Germantown, Maryland. From 2003 to 2006, he was a Research Associate with the University of Maryland. From 2006 to 2008, he was an Assistant Professor with Boise State University, Idaho. He is currently a John and Rebecca Moores Professor with the Electrical and Computer Engineering Department and the Computer Science Department, University of Houston, Texas. He is also a Chair Professor with National Chiao Tung University, China. His research interests include wireless resource allocation and management, wireless communications and networking, game theory, big data analysis, security, and smart grid. He received the NSF Career Award, in 2010, the Fred W. Ellersick Prize of the IEEE Communication Society, in 2011, the Best Paper Award of EURASIP Journal on Advances in Signal Processing, in 2015, the IEEE Leonard G. Abraham Prize in the field of communications systems (best paper award in the IEEE JSAC), in 2016, and several best paper awards in the IEEE conferences. He has served as the IEEE Communications Society Distinguished Lecturer, from 2015 to 2018. He has been a 1\% highly cited researcher, since 2017, according to Web of Science.
\end{IEEEbiography}

\begin{IEEEbiography}[{\includegraphics[width=1in,height=1.25in,clip,keepaspectratio]{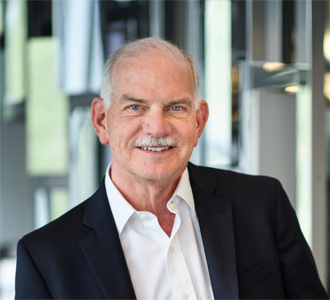}}]
{H. Vincent Poor} (S'72-M'77-SM'82-F'87) received the Ph.D. degree in EECS from Princeton University in 1977.  From 1977 until 1990, he was on the faculty of the University of Illinois at Urbana-Champaign. Since 1990 he has been on the faculty at Princeton, where he is the Michael Henry Strater University Professor of Electrical Engineering. From 2006 until 2016, he served as Dean of Princeton¡¯s School of Engineering and Applied Science. He has also held visiting appointments at several other institutions, including most recently at Berkeley and Cambridge. His research interests are in the areas of information theory, signal processing and machine learning, and their applications in wireless networks, energy systems and related fields. Among his publications in these areas is the forthcoming book Advanced Data Analytics for Power Systems (Cambridge University Press, 2020).

Dr. Poor is a member of the National Academy of Engineering and the National Academy of Sciences, and is a foreign member of the Chinese Academy of Sciences, the Royal Society and other national and international academies. He received the Technical Achievement and Society Awards of the IEEE Signal Processing Society in 2007 and 2011, respectively. Recent recognition of his work includes the 2017 IEEE Alexander Graham Bell Medal, the 2019 ASEE Benjamin Garver Lamme Award, a D.Sc. honoris causa from Syracuse University, awarded in 2017, and a D.Eng. honoris causa from the University of Waterloo, awarded in 2019.
\end{IEEEbiography}

\clearpage
%\newpage
\appendices

\section{Proof of Theorem~\ref{theorem_upper}}\label{appendix_theo1}
From \cite{DBLP:journals/jsac/WangTSLMHC19}, the following lemma presents an upper bound on the loss function in the standard FL.
\begin{lemma}[\cite{DBLP:journals/jsac/WangTSLMHC19}]
An upper bound on the loss function is given by
\begin{equation}\label{lemma_equation}
F(\bar{\boldsymbol{w}}^K)-F(\bar{\boldsymbol{w}}^*)\leq \frac{1}{K\tau(\eta \phi -\frac{\xi h(\tau)}{\tau\varepsilon^2})},
\end{equation}
where
\begin{equation}
\begin{split}
h(x) = \frac{\delta}{L}\left((\eta L+1)^x-1)\right)-\eta\delta x, \quad \varepsilon>0, \\
\phi=\frac{(1-\frac{\eta L}{2})}{\Vert\bar{\boldsymbol{w}}^0-\bar{\boldsymbol{w}}^*\Vert_2}, \quad \left(\eta \phi -\frac{\xi h(\tau)}{\tau\varepsilon^2}\right)>0,
\end{split}
\end{equation}
$\bar{\boldsymbol{w}}^0$ denotes the initial weight, $\bar{\boldsymbol{w}}^*$ denotes the optimal global weight, and $\eta$ denotes the learning rate with $\eta L<1$, respectively.
%and $\xi h(\overline{\tau})$ is derived by $F(\tilde{\boldsymbol{w}(t)})-F(\boldsymbol{w}(t))\leq \xi h(\overline{\tau})$.
\end{lemma}

From (\ref{eq_t_allocation}), we have $\tau = \frac{1}{\alpha} \left(\frac{t^{\mathrm{sum}}}{K }-\beta\right)$. Substituting $\tau$ into  (\ref{lemma_equation}), it yields
\begin{equation}
\begin{split}
F(\bar{\boldsymbol{w}}^{K})-F(\bar{\boldsymbol{w}}^*)&\leq G(K, \alpha, \beta, \eta, \delta, t^{\mathrm{sum}})\\
&=\frac{1}{\gamma \left(\eta\phi-\frac{\frac{\delta\xi K}{L}\left( \lambda^{\frac{\gamma}{K}}-1\right)-\eta\xi\delta \gamma}{\varepsilon^2 \gamma}\right)},
\end{split}
\end{equation}
which concludes the proofs.

\section{Proof of Theorem \ref{theorem_convex}}\label{appendix_theo2}
From (\ref{final_equation}), we define
\begin{equation}\label{g(K)}
g(K) = \frac{1}{G(K)}=\gamma\eta\phi-\frac{\frac{\delta\xi K}{L}\left( \lambda^{\frac{\gamma}{K}}-1\right)-\eta\xi\delta \gamma}{\varepsilon^2},
\end{equation}
where $G(K)>0$ and $g(K)>0$.
Since $g(K)$ is an univariate function, we can optimize $K$ to maximize $g(K)$.
Notice that $\phi,\alpha,\beta,\lambda$ are independent of $K$, and $\gamma$ is a function with respect to $K$. Therefore, we compute the first derivative and second derivative of $\gamma$ with respect to $K$, respectively, as
\begin{equation}\label{equation_derivative}
\begin{split}
\frac{\mathrm{d}\gamma}{\mathrm{d}K}=\frac{-\beta}{\alpha}, \quad
\frac{\mathrm{d}^2\gamma}{\mathrm{d}K^2}=0.
\end{split}
\end{equation}
Then, we have
\begin{equation}\label{g'T}
\begin{split}
\frac{\mathrm{d}g(K)}{\mathrm{d}K}=&\eta\phi \frac{\mathrm{d}\gamma}{\mathrm{d}K}
-\frac{\delta\xi}{\varepsilon^2 L}\left(\lambda^{\frac{\gamma}{K}}-1\right) \\
&+\frac{\eta\delta\xi\frac{\mathrm{d}\gamma}{\mathrm{d}K}}{\varepsilon^2}
-\frac{\delta\xi K}{\varepsilon^2 L}\frac{\mathrm{d}\left(\frac{\gamma}{K}\right)}{\mathrm{d}K}\lambda^{\frac{\gamma}{K}}\ln\lambda,
\end{split}
\end{equation}
and
\begin{equation}\label{eq_g''}
\begin{split}
\frac{\mathrm{d}^2g(K)}{\mathrm{d}K^2}=&-\frac{\delta\xi}{\varepsilon^2 L}\frac{\mathrm{d}\left(\frac{\gamma}{K}\right)}{\mathrm{d}K}\lambda^{\frac{\gamma}{K}}\ln\lambda
-\frac{\delta\xi}{\varepsilon^2 L}\frac{\mathrm{d}\left(\frac{\gamma}{K}\right)}{\mathrm{d}K}\lambda^{\frac{\gamma}{K}}\ln\lambda\\
&-\frac{\delta\xi K}{\varepsilon^2 L}\left[\frac{\mathrm{d}\left(\frac{\gamma}{K}\right)}{\mathrm{d}K}\right]^2\lambda^{\frac{\gamma}{K}}\left(\ln\lambda\right)^2\\
&-\frac{\delta\xi K}{\varepsilon^2 L}\frac{\mathrm{d}^2\left(\frac{\gamma}{K}\right)}{\mathrm{d}K^2}\lambda^{\frac{\gamma}{K}}\ln\lambda.
\end{split}
\end{equation}
%We substitute (\ref{equation_derivative}) into~(\ref{eq_g''}), and obtain
%\begin{equation}
%\begin{split}
%\frac{\mathrm{d}^2g(K)}{\mathrm{d}K^2}=
%-\frac{\delta\xi}{\varepsilon^2 L}\lambda^{\frac{\gamma}{K}}\left(\ln\lambda\right)^2\frac{\left(t^{\mathrm{sum}}\right)^2}{\alpha^2 K^3}<0.
%\end{split}
%\end{equation}
%where $\xi,\delta,\varepsilon,L,T,t^{\mathrm{sum}},\alpha,\gamma>0$ and $\lambda>1$.
Since that $g(K)>0$ and $G(K)=\frac{1}{g(K)}$, we have
\begin{equation}\label{eq_dd}
\frac{\mathrm{d}^2 G(K)}{\mathrm{d}K^2}= \frac{2[\frac{\mathrm{d}g(K)}{\mathrm{d}K}]-\frac{\mathrm{d}g(K)}{\mathrm{d}K} \frac{\mathrm{d}^2 g(K)}{\mathrm{d}K^2} g(K)}{g(K)^3}.
\end{equation}
Substituting (\ref{equation_derivative}), (\ref{g'T}) and (\ref{eq_g''}) into (\ref{eq_dd}), it yields
\begin{equation}
\begin{split}\label{eq_dd}
&\frac{\mathrm{d}^2 G(K)}{\mathrm{d}K^2}= \nonumber \\
&\frac{2\eta^2 \phi^2 \beta^2 L^2 \varepsilon^4 \frac{\beta}{\alpha} + \left(2 \alpha^2 \delta^2 \xi^2 + \frac{(t^{\mathrm{sum}})^2 \delta^3 \xi^3}{\varepsilon^2 L K^2}\right)(\lambda^{\frac{\gamma}{K}}-1)^2}{g(K)^3} > 0.
\end{split}
\end{equation}

Thus, we can prove that $G(K)$ is convex and has its own minimum value.
\section{Proof of Theorem \ref{theorem_closed}} \label{appendix_theo3}
Let $\varepsilon^2=\frac{\delta\xi}{\phi}$ and $\frac{\eta L \gamma}{K}\ll1$. We first have
\begin{equation}\label{temp}
\begin{split}
\lambda^{\frac{\gamma}{K}}&=(1+\eta L)^{\frac{\gamma}{K}}\\
&=1+\eta L\frac{\gamma}{K}+\frac{\frac{\gamma}{K}\left(\frac{\gamma}{K}-1\right)}{2}(\eta L)^2+O\left((\eta L)^3\right)\\
&<1+2\eta L\frac{\gamma}{K},
\end{split}
\end{equation}
where
\begin{equation}
\frac{\gamma}{K}=\tau\geq 1, \quad \eta L<1.
\end{equation}
Using~(\ref{temp}), we obtain
\begin{equation}
\gamma\eta\phi-\frac{\frac{\delta\xi K}{L}\left( \lambda^{\frac{\gamma}{K}}-1\right)-\eta\xi\delta \gamma}{\varepsilon^2}>0.
\end{equation}
Then, we approximate $\lambda^{\frac{\gamma}{K}}$ as a quadratic term with Taylor expansion:
\begin{equation}\label{taylor_expansion}
\lambda^{\frac{\gamma}{K}}=\left(1+\eta L\right)^{\frac{\gamma}{K}}=1+\eta L\frac{\gamma}{K}+\frac{\frac{\gamma}{K}(\frac{\gamma}{K}-1)}{2}\eta^2 L^2.
\end{equation}
Thus, $g(K)$ can be written as
\begin{equation}
\begin{split}
g(K) =\eta\phi\gamma\left[1-\frac{\eta L}{2}\left(\frac{\gamma}{K}-1\right)\right].
\end{split}
\end{equation}
To solve the convex problem, we let $\left(\frac{\mathrm{d}g(K)}{\mathrm{d}K}\vert_{K=K^*}\right)=0$, i.e.,
\begin{equation}
\begin{split}
&\left(\frac{\mathrm{d}g(K)}{\mathrm{d}K}\vert_{K=K^*}\right)=-\beta+\frac{1}{2}\eta L \beta\left[\frac{t^{\mathrm{sum}}}{\alpha K^*}-\frac{\beta}{\alpha}-1\right]\\
&+t^{\mathrm{sum}}\left(\frac{t^{\mathrm{sum}}}{2\alpha {(K^*)}^2}\eta L\right)
-K^*\beta\left(\frac{t^{\mathrm{sum}}}{2\alpha {(K^*)}^2}\eta L\right)\\
&=\frac{\eta L \left(t^{\mathrm{sum}}\right)^2}{2\alpha {(K^*)}^2}-\left(\beta+\frac{\eta L \beta^2}{2\alpha}+\frac{\eta L \beta}{2}\right)=0.
\end{split}
\end{equation}
Finally, we have
\begin{equation}
\begin{aligned}
%\alpha\frac{{t^{\mathrm{sum}}}^2}{{T^*}^2}=\alpha\beta^2+\beta+\frac{2\beta}{\eta L},\\
K^* = \frac{t^{\mathrm{sum}}}{\sqrt{\frac{2 \alpha\beta}{\eta L}+\alpha\beta+\beta^2}}.
\end{aligned}
\end{equation}
This completes the proof.

\section{Proof of Corollary \ref{delta_remark}} \label{appendix_corodelta}
Without approximation of (\ref{taylor_expansion}), we first let $\left(\frac{\mathrm{d}g(K)}{\mathrm{d}K}\vert_{K=K^*}\right)=0$, i.e.,
\begin{equation}\label{dg_T*}
\begin{aligned}
\lambda^{\frac{\gamma}{K^*}}\left(\frac{t^{\mathrm{sum}}\ln\lambda}{\alpha K^*}-1\right)-\left(\frac{\eta\phi\varepsilon^2 L}{\delta\xi\alpha}+\frac{\eta L\beta}{\alpha}-1\right)=0,
\end{aligned}
\end{equation}
For simplicity, we let
\begin{equation}\label{dg'_T*}
\begin{aligned}
p(x)=x\lambda^{x}\ln\lambda -\lambda^{x}-\Omega=0,
\end{aligned}
\end{equation}
where
\begin{equation}\label{omega}
x=\frac{t^{\mathrm{sum}}}{\alpha K^*}, \quad \Omega=\lambda^{\frac{\beta}{\alpha}}\left[\frac{\eta\phi\varepsilon^2 L}{\delta\xi\alpha}+\frac{\eta L\beta}{\alpha}-1\right].
\end{equation}
Then, the first derivative of $p(x)$ is given by
\begin{equation}
\frac{\mathrm{d}p(x)}{\mathrm{d}x} =\lambda^x(\ln\lambda)^2+\lambda^x\ln\lambda-\lambda^x\ln\lambda=\lambda^x(\ln\lambda)^2>0.
\end{equation}
Notice that $\Omega$ in (\ref{dg'_T*}) is a decreasing function with respect to $\delta$, $p(x)$ is an increasing function with respect to $x$, and
\begin{equation}
K^*=\frac{t^{\mathrm{sum}}}{\alpha x}
\end{equation}
is a decreasing function function with respect to $x$, respectively. Thus, the solution $x$ of~(\ref{dg'_T*}) drops as $\delta$ grows. Finally, we conclude that $K^*$ increases as $\delta$ rises.

\section{Proof of Corollary \ref{eta_remark}} \label{appendix_coroeta}
From~(\ref{omega}), we know that $\Omega$ increases as the learning rate $\eta$ rises, which leads to larger $x$. Thus, from the proof of Corollary \ref{delta_remark}, $K^*$ descends as $\eta$ ascends. Then the derivative of the function $g(\cdot)$ with respect to $\eta$ is
\begin{equation}\label{d_eta}
\frac{\mathrm{d}g}{\mathrm{d}\eta} = \gamma \left(\phi+\frac{\delta\xi}{\varepsilon^2}\right)>0,
\end{equation}
where $\eta L<1$. It indicates that the loss function decreases as rate $\eta$ increases if $\eta L<1$. However, the condition $\eta L<1$ is not satisfied when $\eta$ is sufficiently large. In this case, $g(\cdot)$ is not an increasing function with respect to $\eta$, resulting in larger loss function. This completes the proof.

\section{Proof of Theorem \ref{theorem_lazyupper}} \label{appendix_theolazy}
Define the model weights $\tilde{\boldsymbol{w}}^{k}$ of lazy clients as
\begin{equation}
\begin{split}
\tilde{\boldsymbol{w}}^{k}&= \frac{1}{N}\left(\sum_{j=1}^N \boldsymbol{w}_j^k-\sum_{j=1}^M \boldsymbol{w}_j^k+\sum_{j'=1}^M \boldsymbol{w}_{j'}^k+\sum_{i=1}^M \mathbf{n}_i\right)\\
&=\bar{\boldsymbol{w}}^{k}-\frac{1}{N}\left(\sum_{j=1}^M \boldsymbol{w}_j^k-\sum_{j'=1}^M \boldsymbol{w}_{j'}^k-\sum_{i=1}^M \mathbf{n}_i\right),
\end{split}
\end{equation}
where $\boldsymbol{w}_{j'}^k$ is the model parameters that are plagiarized by lazy clients.

Since $F(\bar{\boldsymbol{w}}^k)$ is $\xi$-Lipschitz, the proof of \textbf{Lemma 1} in \cite{DBLP:journals/jsac/WangTSLMHC19} has shown that
%\begin{equation}
\begin{align}
\nonumber &\mathbb{E}\{F(\tilde{\boldsymbol{w}}^{k})\}-F(\bar{\boldsymbol{w}}^*)\\
\nonumber =&\mathbb{E}\{F(\tilde{\boldsymbol{w}}^{k})\}-F(\bar{\boldsymbol{w}}^k)+(F(\bar{\boldsymbol{w}}^k)-F(\bar{\boldsymbol{w}}^*))\\
\nonumber \leq&\xi\mathbb{E}\{\Vert(\tilde{\boldsymbol{w}}^{k}-\bar{\boldsymbol{w}}^k)\Vert_2\}
+(F(\bar{\boldsymbol{w}}^k)-F(\bar{\boldsymbol{w}}^*))\\
\nonumber \leq & \frac{\xi}{N}\mathbb{E}\{\Vert\sum_{j'=1}^M \boldsymbol{w}_{j'}^k-\sum_{j=1}^M \boldsymbol{w}_j^k\Vert_2\}
+\frac{\xi}{N}\Vert\mathbb{E}\{\sum_{j=1}^M \mathbf{n}_i\}\Vert_2+\xi h(\tau)\\
\leq & \frac{M}{N}\xi\theta+\frac{\xi}{N}\sqrt{\sum_{i=1}^M \sigma_i^2}
+\frac{\delta\xi K}{L}\left( \lambda^{\frac{\gamma}{K}}-1\right)-\eta\xi\delta \gamma.
\end{align}
%\end{equation}
Therefore, the upper bound can be expressed as~\cite{DBLP:journals/jsac/WangTSLMHC19}
\begin{equation}\label{lazy_equation_diff}
\begin{split}
&F(\tilde{\boldsymbol{w}}^{K})-F(\bar{\boldsymbol{w}}^*)\\
&\leq\frac{1}{\gamma \left(\eta\phi-\frac{\frac{\delta\xi K}{L}\left( \lambda^{\frac{\gamma}{K}}-1\right)-\eta\xi\delta \gamma+K\xi \frac{M}{N}\theta+\frac{K\xi}{N}\sqrt{\sum_{i=1}^M \sigma_i^2}}{\varepsilon^2 \gamma}\right)}.
\end{split}
\end{equation}
In addition, plugging (\ref{final_equation}) into (\ref{lazy_equation_diff}), we have
\begin{equation}\label{lazy_equation_theta}
\begin{split}
&\frac{1}{F(\tilde{\boldsymbol{w}}^{K})-F(\bar{\boldsymbol{w}}^*)}\\
&=\frac{1}{F(\bar{\boldsymbol{w}}^{K})
-F(\bar{\boldsymbol{w}}^*)}
+\frac{K\xi}{\varepsilon^2 N}\left(M \theta + \sqrt{\sum^M_{i=1} \sigma_i^2}\right).
\end{split}
\end{equation}
From (\ref{lazy_equation_theta}), we further have
\begin{equation}
\begin{split}
\theta=\frac{\varepsilon^2 N}{K\xi M}\left(\frac{1}{F(\tilde{\boldsymbol{w}}^{K})-F(\bar{\boldsymbol{w}}^*)}
-\frac{1}{G(K)}\right)-\frac{\sqrt{\sum^M_{i=1} \sigma_i^2}}{M}.
\end{split}
\end{equation}

%Indeed, $\theta$ is a function of $T$ and $\tau$, while $\theta(T,\tau)$ is a reduction function of either $T$ or $\tau$.

If each lazy client adds the Gaussian noise with the same variance to its plagiarized model, $\sum_{i=1}^M \frac{N_i}{\sigma_i}$ is a chi-square distribution with $M$ degrees of freedom (i.e., $\chi^2(M)=\sum^M_{i=1}X_i, \forall X_i \sim \mathcal{N}(0,1) $). Given the mean value $\mathbb{E}\{\chi^2(M)\}=M$, we have
\begin{equation}
\left\Vert\mathbb{E}\left\{\sum_{i=1}^M \mathbf{n}_i\right\}\right\Vert_2= \sqrt{M}\sigma^2.
\end{equation}
The upper bound in~(\ref{lazy_equation_diff}) can be written as
\begin{equation}
\begin{split}
&F(\tilde{\boldsymbol{w}}^{K})-F(\bar{\boldsymbol{w}}^*) \leq \tilde{G}(K, \alpha, \beta, \eta, \delta, t^{\mathrm{sum}}, \theta, \sigma^2)\\
&= \frac{1}{\gamma \left(\eta\phi-\frac{\frac{\delta\xi K}{L}\left( \lambda^{\frac{\gamma}{K}}-1\right)-\eta\xi\delta \gamma+K\xi\frac{M}{N}\theta+K\xi\frac{\sqrt{M}}{N}\sigma^2}{\varepsilon^2 \gamma}\right)}.
\end{split}
\end{equation}
This completes the proof.

\section{Proof of Corollary \ref{pro_lazy_sigma}} \label{appendix_corolazy}
From the definition of $g(K)$ in (\ref{g(K)}), we let
\begin{equation}
L(K)=g(K)-\frac{K}{\varepsilon^2}\left(\frac{\xi M}{N}\theta+\frac{\xi \sqrt{M}}{N}\sigma^2\right).
\end{equation}
As such, $\frac{1}{L(K)}=\tilde{G}(K)$, which represents the loss function of BLADE-FL with lazy clients.

Since we have
\begin{equation}
\frac{\mathrm{d}L(K)}{\mathrm{d}K}=\frac{\mathrm{d}g(K)}{\mathrm{d}K}-\frac{1}{\varepsilon^2}\left(\frac{\xi M}{N}\theta+\frac{\xi \sqrt{M}}{N}\sigma^2\right),
\end{equation}
and
\begin{equation}
\frac{\mathrm{d}^2L(K)}{\mathrm{d}K^2}=\frac{\mathrm{d}^2g(K)}{\mathrm{d}K^2},
\end{equation}
we obtain that $L(K)$ is still convex with respect to $K$.

Furthermore, we let $\varepsilon^2=\frac{\delta\xi}{\phi}$. Plugging this into $\left(\frac{\mathrm{d}L(K)}{\mathrm{d}K}\vert_{K=K^*}\right)=0$, we have
\begin{equation}\label{L(T)}
\begin{split}
&\lambda^{\frac{\gamma}{K^*}}\left(\frac{t^{\mathrm{sum}}\ln\lambda}{\alpha K^*}-1\right)\\
&-\left(\frac{\eta\phi\varepsilon^2 L}{\delta\xi\alpha}+\frac{\eta L\beta}{\alpha}-1+\frac{L}{\delta}\left(\frac{M}{N}\theta
+\frac{\sqrt{M}}{N}\sigma\right)\right)
=0.
\end{split}
\end{equation}

Then we let
\begin{equation}
x=\frac{ t^{\mathrm{sum}}}{\alpha K^*},
\end{equation}
and express (\ref{L(T)}) as
\begin{equation}\label{L'(x)}
\begin{split}
&L'(x)=x\lambda^x\ln\lambda-\lambda^x\\
&-\left[\lambda^{\frac{\beta}{\alpha}}\left(\frac{\eta\phi\varepsilon^2 L}{\delta\xi\alpha}+\frac{\eta L\beta}{\alpha}+\frac{L}{\delta}\left(\frac{M}{N}\theta+\frac{\sqrt{M}}{N}\sigma^2\right)-1\right)\right]=0.
\end{split}
\end{equation}
We notice that
\begin{equation}
\frac{\mathrm{d}L'(x)}{\mathrm{d}x}=x\lambda^x(\ln\lambda)^2>0.
\end{equation}
Thus $L'(x)$ is an increasing function with respect to $x$. Let
\begin{equation}
\Omega'=\lambda^{\frac{\beta}{\alpha}}\left[\frac{\eta\phi\varepsilon^2 L}{\delta\xi\alpha}+\frac{\eta L\beta}{\alpha}+\frac{L}{\delta}\left(\frac{M}{N}\theta+\frac{\sqrt{M}}{N}\sigma^2\right)-1\right].
\end{equation}
Thus (\ref{L'(x)}) can be rewritten as
\begin{equation}
\begin{aligned}
x\lambda^x\ln\lambda-\lambda^x=\Omega',
\end{aligned}
\end{equation}
where $\Omega'$ grows as either $\frac{M}{N}$ or $\sigma^2$ increases, $x$ goes up as $\Omega'$ grows, and $K^*$ declines as $x$ goes up. Finally, $K^*$ that minimizes $\tilde{G}(\cdot)$ in (\ref{eq_lazy_sigma}) decreases as either $\frac{M}{N}$ or $\sigma^2$ grows. This concludes the proof.

\end{document}